\documentclass[letterpaper]{article} 
\usepackage[preprint]{aaai2027}  
\usepackage[hyphens]{url}  
\usepackage{graphicx} 
\usepackage{natbib}  
\usepackage{caption} 
\usepackage{algorithm}
\usepackage{algorithmic}

\usepackage{newfloat}
\usepackage{listings}
\DeclareCaptionStyle{ruled}{labelfont=normalfont,labelsep=colon,strut=off} 
\floatstyle{ruled}
\newfloat{listing}{tb}{lst}{}
\floatname{listing}{Listing}

\usepackage{booktabs}

\usepackage{amsmath}
\usepackage{amssymb}

\usepackage{graphicx}

\usepackage{textcomp}

\newcommand{\res}[2]{%
  #1\,\textpm\,#2%
}
\newcommand{\best}[2]{%
  \textbf{#1\,\textpm\,#2}%
}
\newcommand{\second}[2]{%
  \underline{#1\,\textpm\,#2}%
}

\newcommand{\bestmean}[1]{\textbf{#1}}
\newcommand{\secondmean}[1]{\underline{#1}}

\usepackage{tabularx}
\usepackage{array}
\usepackage{multirow}

\title{FeatureHospital: A Skill-Driven Multi-Agent Framework for Automated Algorithm Customization in Multi-View Multi-Label Feature Selection}
\author{
Junxuan Li\textsuperscript{\rm 1} \quad
Zhiqi Chen\textsuperscript{\rm 1} \quad
Yuzhou Liu\textsuperscript{\rm 1*} \quad
Peng Zhang\textsuperscript{\rm 1} \quad
Huaxiao Liu\textsuperscript{\rm 1}
}

\affiliations{
\textsuperscript{\rm 1}College of Computer Science and Technology, Jilin University, Jilin, China\\
\textsuperscript{*}Corresponding author.
}

\usepackage{algorithm}
\usepackage{algorithmic}
\usepackage{amsmath}
\usepackage{amssymb}
\usepackage{multirow}

\usepackage{newfloat}
\usepackage{listings}
\DeclareCaptionStyle{ruled}{labelfont=normalfont,labelsep=colon,strut=off} 
\floatstyle{ruled}
\newfloat{listing}{tb}{lst}{}
\floatname{listing}{Listing}

\usepackage{booktabs}

\usepackage{booktabs}

\newcommand{\headleft}[1]{%
  \shortstack[l]{\textbf{#1}\\\phantom{\textbf{27B}}}%
}
\newcommand{\headone}[1]{%
  \shortstack[c]{\textbf{#1}\\\phantom{\textbf{27B}}}%
}
\newcommand{\headtwo}[2]{%
  \shortstack[c]{\textbf{#1}\\\textbf{#2}}%
}

\newcommand{\val}[2]{%
  \mbox{#1 \(\pm\) #2}%
}
\newcommand{\avgval}[1]{%
  \mbox{\vphantom{0.0000 \(\pm\) 0.0000}#1}%
}

\newcommand{\avgbest}[1]{%
  \textbf{\avgval{#1}}%
}

\newcommand{\avgsecond}[1]{%
  \underline{\avgval{#1}}%
}

\newcommand{\avglabel}{%
  \textbf{\vphantom{0.0000 \(\pm\) 0.0000}Average}%
}

\newcommand{\rowstrut}{%
  \rule[-0.65ex]{0pt}{2.55ex}%
}

\providecommand{\FHCode}[1]{\texttt{\detokenize{#1}}}

\usepackage{longtable}
\usepackage{booktabs}
\usepackage{multirow}
\usepackage[most]{tcolorbox}
\usepackage{xcolor}
\usepackage{makecell}

\usepackage{multicol}

\newtcolorbox{casebox}[1]{
breakable,
enhanced,
colback=gray!10,
colframe=gray!60,
title=#1,
fonttitle=\bfseries,
coltitle=black,
boxrule=0.8pt,
arc=2mm,
left=2mm,
right=2mm,
top=1mm,
bottom=1mm
}
\usepackage{graphicx}

\begin{document}

\maketitle

\begin{abstract}
Multi-view multi-label feature selection aims to identify a compact and informative feature subset from heterogeneous views while preserving discriminative information for multiple labels. Existing methods are generally developed from specific modeling perspectives and incorporate mechanisms tailored to particular data characteristics. Designing suitable feature selection algorithms across datasets with diverse and heterogeneous characteristics still relies heavily on expert knowledge and substantial manual effort, imposing considerable time and labor costs that severely hinder the practical adoption of feature selection. To address this problem, we propose FeatureHospital, a Skill-driven multi-agent framework for automated multi-view multi-label feature selection algorithm design. FeatureHospital first diagnoses the target dataset to identify its feature selection issues. Based on the diagnosis, specialist agents equipped with domain Skills then prescribe corresponding optimization strategies and Loss terms for different issues. After that, the resulting prescriptions are reconciled to remove overlaps and resolve conflicts before being integrated into a compact dataset-specific objective. Finally, the constructed objective is optimized to select the final feature subset. Experimental results demonstrate that FeatureHospital can construct effective feature selection algorithms for different datasets based on their individual characteristics.
\end{abstract}
\begin{figure*}[t]
  \centering
  \includegraphics[width=\linewidth]{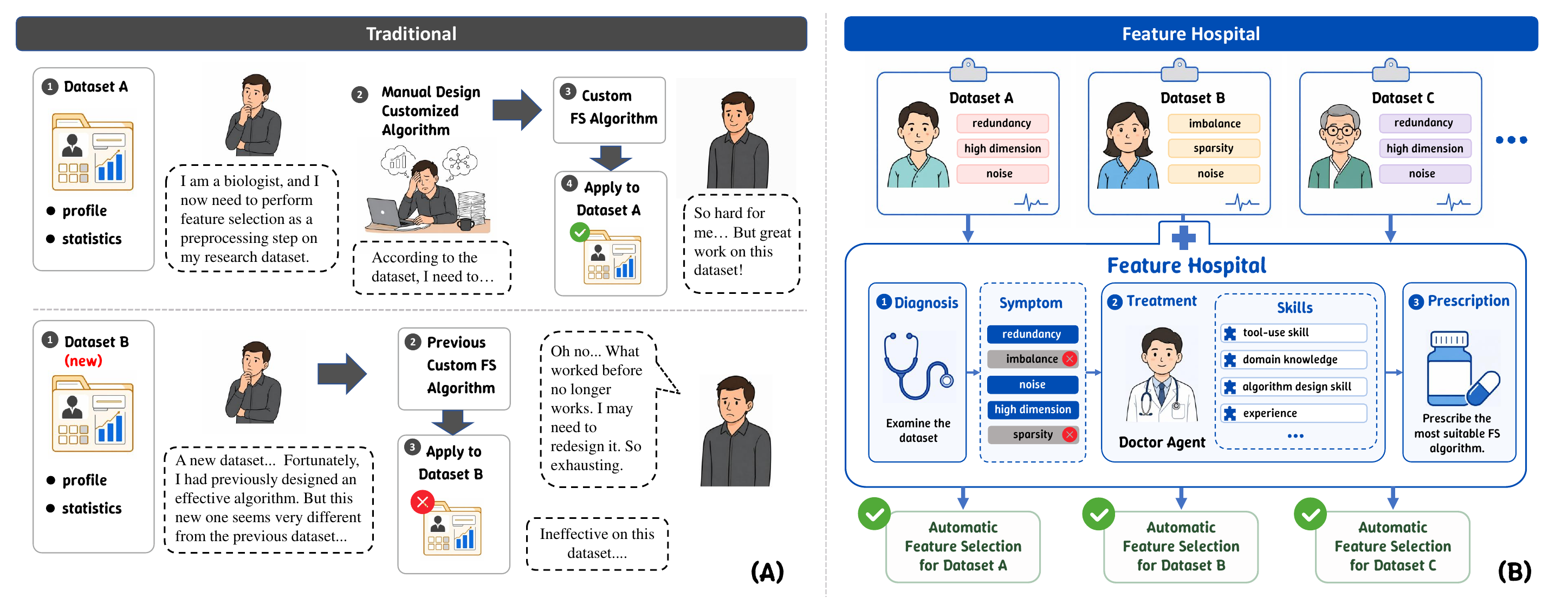}
  \caption{
Conceptual comparison between conventional feature
selection algorithm design and FeatureHospital.
(A) In the conventional workflow, researchers manually design a
feature selection algorithm for a particular dataset, and the same
algorithm may become ineffective when applied to a new dataset,
requiring repeated analysis and redesign.
(B) FeatureHospital treats each dataset as a patient, diagnoses its
feature selection issues, and employs Skill-equipped agents to
automatically construct a dataset-specific feature selection algorithm.
}
  \label{fig:method}
\end{figure*}

\section{Introduction}

In real-world scientific applications, data instances are frequently described by multiple heterogeneous feature sets from distinct views and associated with multiple semantic labels~\citep{yan2022multiview}. Data with this form are commonly referred to as Multi-View Multi-Label (MVML) data. However, such data also introduce complex challenges, including view-quality imbalance~\citep{liu2023incomplete}, feature redundancy~\citep{han2024feature}, missing labels~\citep{wen2023deep}, label imbalance~\citep{charte2015addressing}, and label dependency~\citep{liu2023multi}. 
To reduce data complexity and enhance representation quality, multi-view multi-label feature selection aims to select a compact and informative feature subset from multiple views while preserving discriminative information for multiple labels. Existing multi-view multi-label feature selection methods typically address these challenges from specific perspectives, such as exploiting view complementarity, label correlations, or feature redundancy structures, and have achieved promising performance~\cite{hao2024dhli,hao2024grafs,zhang2020msfs,gonzalez2020enm}.

Despite these advances, current feature selection methods still face practical challenges in real-world research applications. As illustrated in Figure~\ref{fig:method}(A), researchers often use feature selection as a practical tool rather than study it as their primary research topic. This leaves them with a recurring algorithm-design burden: (1) selecting or designing a suitable feature selection method is difficult because it requires both an understanding of the dataset characteristics and specialized knowledge of feature selection algorithms. (2) This difficulty is not a one-time burden. When a new dataset has underlying characteristics that do not align with the assumptions of the previously adopted algorithm, the researcher must repeat the same unfamiliar process of method selection and algorithm design. These challenges consume substantial time and effort that could otherwise be devoted to the primary scientific work of researcher. 

Recent advances in LLM-based agents provide a promising foundation for automated algorithm design~\citep{jeong2024llm}. Once equipped with reusable procedural Skills that encapsulate domain knowledge and structured tool-use procedures, general-purpose agents can act as domain experts~\citep{pan2026anything2skill}. This raises a natural question: can a Skill-driven multi-agent system automatically design a feature selection algorithm tailored to the target dataset in the manner of a human expert?

In this work, we propose FeatureHospital, a Skill-driven multi-agent framework for automated modeling in multi-view multi-label feature selection. Figure~\ref{fig:method}(B) provides an overview of the motivation and workflow of FeatureHospital. Instead of relying on researchers to manually analyze each dataset, design optimization objectives, FeatureHospital aims to automate the entire process from dataset diagnosis to algorithm construction, thereby generating dataset-specific feature selection strategies without repeated human intervention. Specifically, we treat each dataset as a ``patient'' and its intrinsic data characteristics as ``symptoms'', such as label imbalance, feature redundancy, view-quality imbalance, and feature-budget pressure. The system first diagnoses the dataset to identify its major issues. Then, specialist doctors equipped with domain Skills prescribe corresponding optimization strategies and loss terms for different problems. These local treatments are combined into a unified prescription for the entire dataset. Finally, the system automatically constructs the optimization objective from this prescription and performs training and feature selection.

In summary, our main contributions are as follows:

\begin{itemize}

\item We propose FeatureHospital, a Skill-driven multi-agent framework for automated multi-view multi-label feature selection algorithm design. Equipped with reusable domain Skills, specialized agents collaboratively analyze the target dataset and design a feature selection algorithm tailored to its characteristics.

\item We develop an automated feature selection algorithm design
paradigm that connects dataset analysis with objective construction.
FeatureHospital identifies potential dataset problems, formulates a problem-specific optimization term for each identified problem, and integrates these terms into a unified
dataset-specific feature selection objective.

\item We conduct experiments on seven multi-view multi-label datasets and compare FeatureHospital with seven representative multi-view multi-label feature selection methods. The results show that the algorithms automatically generated by FeatureHospital achieve competitive performance.
\end{itemize}

\begin{figure*}[t]
  \centering
  \includegraphics[width=0.9\linewidth]{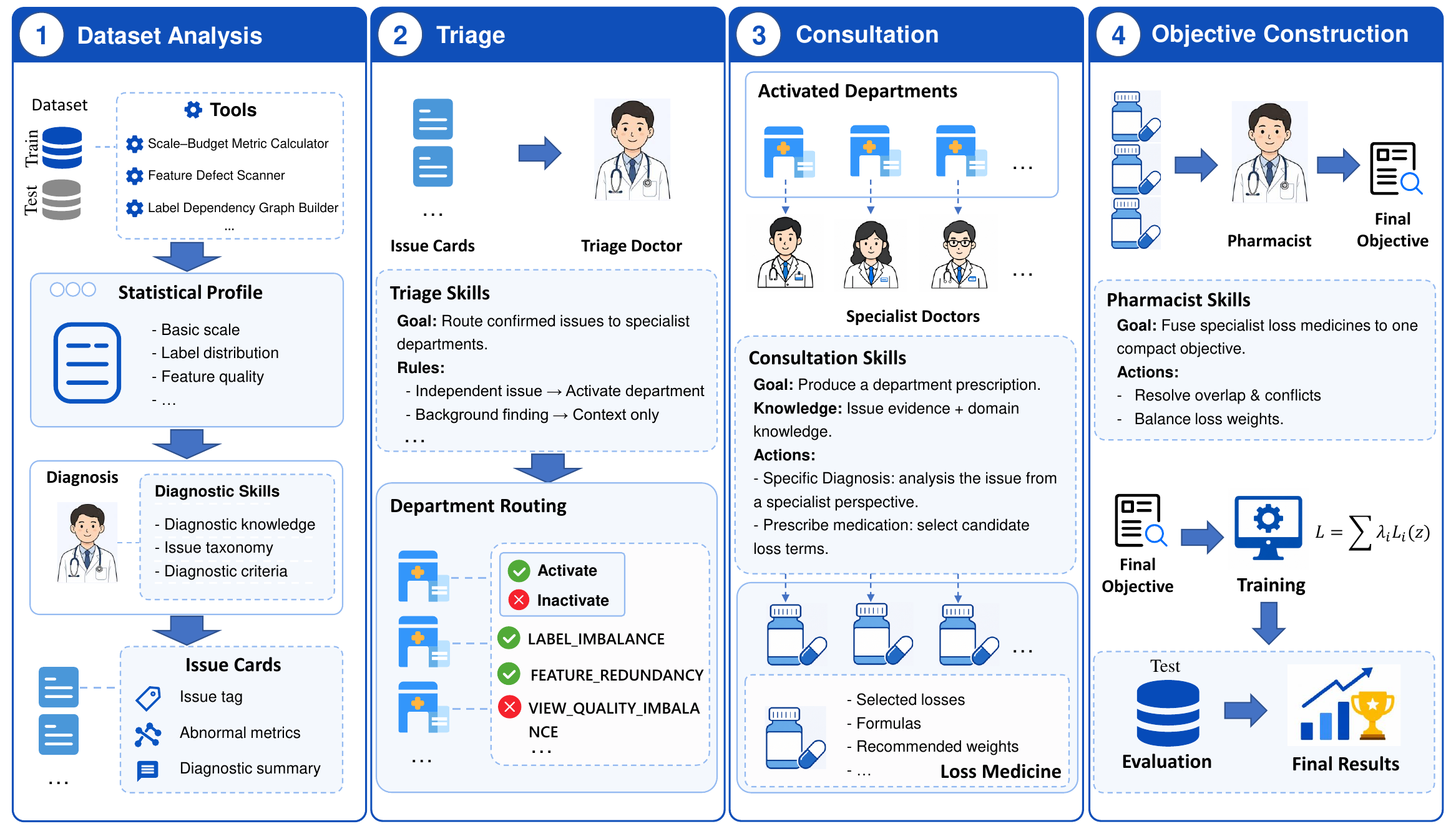}
  \caption{
Overview of FeatureHospital.
The framework proceeds in four stages:
(1) analyze the target dataset and identify potential problems affecting
feature selection;
(2) categorize the identified problems and assign them to suitable
specialist agents;
(3) select corresponding loss functions for each problem;
(4) remove redundant or conflicting loss functions, balance their
contributions, and combine them into a dataset-specific objective; and finally optimize the constructed objective to select features and evaluate
the resulting feature subset.
}
  \label{fig:framework}
\end{figure*}

\section{Related Work}

\subsection{Multi-View Multi-Label Feature Selection}

Multi-view multi-label learning exploits multiple heterogeneous views
and multiple labels to capture rich and complementary semantic
information~\citep{yan2022multiview}. However, the resulting
representations also increase data complexity, making effective data
analysis more challenging~\citep{hao2024dhli}. Multi-view multi-label
feature selection therefore aims to retain a compact and discriminative
feature subset from such data. Existing methods address different data
structures from specific modeling perspectives. Some methods exploit
view-specific or hybrid label information
~\citep{hao2024i2vslc,hao2024dhli}, while others construct informative
cross-view representations through global-view reconstruction or
embedded feature fusion~\citep{hao2024grafs,hao2025ef2fs}. Sparse
learning has also been used to identify informative features
~\citep{zhang2020msfs}.

Although these methods have achieved promising performance, their
objective structures are generally fixed once designed. Since different
datasets may present different combinations of feature selection
problems, determining which characteristics should be modeled and
translating them into a suitable objective still require substantial
domain knowledge and manual algorithm-design effort.

\subsection{Skill-Driven Multi-Agent Systems}

LLM-based multi-agent systems have been increasingly applied to complex
tasks by assigning agents complementary roles and coordinating their
interactions~\citep{fourney2024magentic,hong2024metagpt,su2025many}.
However, general-purpose agents may lack the specialized knowledge and
procedural experience required in different domains. Skill-based
approaches provide a lightweight means of specialization by
encapsulating reusable capabilities that can be retrieved and composed
without retraining the underlying model
~\citep{wang2023voyager,zheng2025skillweaver}. Existing studies
generally define a Skill as a callable module that packages domain or
procedural knowledge, executable tools or code, and
operating rules~\citep{jiang2026sok,xu2026agent}. Such Skills enable
general-purpose agents to acquire domain-specific capabilities and
perform tasks in a manner similar to human experts. Nevertheless, the
use of Skill-driven multi-agent systems for feature selection remains
largely unexplored.

Our work bridges these two research directions by proposing
FeatureHospital, a Skill-driven multi-agent framework for MVML feature
selection. FeatureHospital equips specialized agents with feature
selection Skills to diagnose dataset-specific issues, translate them
into corresponding optimization strategies, and collaboratively design
feature selection algorithms for different datasets. In this way, it
reduces the dependence of MVML feature selection algorithm design on
specialized expertise and extensive manual effort.

\section{Methodology}
Figure~\ref{fig:framework} illustrates the overall workflow of FeatureHospital.
\subsection{Problem Formulation}

Let $\mathcal{D}=(\mathcal{X},Y)$ denote a MVML dataset, where
\begin{equation}
\begin{aligned}
\mathcal{X}
&=
\left\{X^{(v)}\right\}_{v=1}^{V},
\quad
X^{(v)}
\in
\mathbb{R}^{n \times d_v}, \\
Y
&\in
\{0,1\}^{n \times l}.
\end{aligned}
\end{equation}

Here, $V$ is the number of views, $n$ is the number of instances, $d_v$ is the number of features in the $v$-th view, and $l$ is the number of labels. The total number of features is $d=\sum_{v=1}^{V}d_v$.

For each experimental run, $\mathcal{D}$ is divided into two
disjoint subsets:
\begin{equation}
\mathcal{D}
=
\mathcal{D}_{\mathrm{tr}}
\cup
\mathcal{D}_{\mathrm{te}},
\qquad
\mathcal{D}_{\mathrm{tr}}
\cap
\mathcal{D}_{\mathrm{te}}
=
\varnothing,
\end{equation}
where
$\mathcal{D}_{\mathrm{tr}}
=(\mathcal{X}_{\mathrm{tr}},Y_{\mathrm{tr}})$
and
$\mathcal{D}_{\mathrm{te}}
=(\mathcal{X}_{\mathrm{te}},Y_{\mathrm{te}})$
denote the training and test partitions, respectively.

Conventional embedded feature selection methods generally optimize a predefined objective:
\begin{equation}
\omega^{*}
=
\arg\min_{\omega}
\mathcal{L}_{\mathrm{fixed}}
\left(
\mathcal{X}_{tr},Y_{tr};\omega,\theta
\right),
\end{equation}
where $\omega$ denotes the trainable model parameters, and $\theta$ denotes the hyperparameters. During training, $\omega$ is optimized according to the input data. However, the functional form and objective-term composition of $\mathcal{L}_{\mathrm{fixed}}$ are predefined during algorithm design and remain unchanged across datasets.

In contrast, FeatureHospital does not apply the same predefined objective to all datasets. Instead, it constructs a dataset-specific feature selection objective according to the characteristics of the target dataset without requiring manual objective design:
\begin{equation}
\mathcal{L}_{\mathcal{D}}
=
F_{\mathrm{hospital}}
\left(
\mathcal{D}_{tr}
\right).
\end{equation}

The trainable parameters are then optimized with respect to the
constructed objective:
\begin{equation}
\omega^{*}
=
\arg\min_{\omega}
\mathcal{L}_{\mathcal{D}}
\left(
\mathcal{X}_{tr},Y_{tr};\omega,\theta
\right).
\end{equation}

Therefore, different datasets may receive different objective structures and configurations according to their respective characteristics.








\subsection{Dataset Analysis}

Given a target dataset $\mathcal{D}=(\mathcal{X},Y)$, we first partition the target dataset $D$ into two disjoint subsets: train subset $\mathcal{D}_{tr}=(\mathcal{X}_{tr},Y_{tr})$ and test subset $\mathcal{D}_{te}=(\mathcal{X}_{te},Y_{te})$. FeatureHospital has access only to $\mathcal{D}_{\mathrm{tr}}$ before final algorithm evaluation. FeatureHospital first employs a set of executable analysis tools to construct its Statistical Profile:
\begin{equation}
P_{\mathcal{D}}
=
F_{\mathrm{analysis}}
\left(
\mathcal{X}_{tr},Y_{tr};\mathcal{T}_{\mathrm{analysis}}
\right),
\end{equation}
where $\mathcal{T}_{\mathrm{analysis}}$ denotes the tools used for dataset analysis. The resulting Statistical Profile $P_{\mathcal{D}}$ describes general characteristics relevant to feature selection, including data scale and feature budget, label distribution, feature quality, feature redundancy, label dependency, view heterogeneity, etc. The complete specification of $\mathcal{T}_{\mathrm{analysis}}$ and the construction of $P_{\mathcal{D}}$ are provided in Appendix~C.

The Diagnosis Agent subsequently interprets the Statistical Profile using Diagnostic Skills:
\begin{equation}
\mathcal{I}_{\mathcal{D}}
=
F_{\mathrm{diagnosis}}
\left(
P_{\mathcal{D}};
\mathcal{K}_{\mathrm{diagnosis}}
\right),
\end{equation}
where $\mathcal{K}_{\mathrm{diagnosis}}$ contains diagnostic knowledge, an issue taxonomy, and diagnostic criteria. The output $\mathcal{I}_{\mathcal{D}}$ is a collection of structured Issue Cards. Each Issue Card records an identified issue, the abnormal metrics that support it, and a diagnostic summary. 

\subsection{Triage}

The triage stage assigns the issues recorded in the Issue Cards to one
or more predefined problem categories, termed specialist
\emph{Departments} in FeatureHospital. Each Department specifies the
scope of feature selection problems it handles and provides a
specialized context for subsequent algorithm design. Guided by the
Triage Skill $\mathcal{K}_{\mathrm{triage}}$, the Triage Doctor
compares each issue with the responsibility scopes of the Departments,
activates the relevant Departments, and routes the issue accordingly.
Unmatched issues are retained as contextual information.

Formally, the triage process is defined as
\begin{equation}
\left(
\mathcal{H}_{\mathcal{D}}^{+},
\mathcal{R}_{\mathcal{D}}
\right)
=
F_{\mathrm{triage}}
\left(
\mathcal{I}_{\mathcal{D}},
\mathcal{H};
\mathcal{K}_{\mathrm{triage}}
\right),
\end{equation}
where $\mathcal{H}$ is the predefined Department set,
$\mathcal{H}_{\mathcal{D}}^{+}
=\{H_i^{+}\}_{i=1}^{N_{\mathcal{D}}}\subseteq\mathcal{H}$
contains the $N_{\mathcal{D}}$ Departments activated for dataset
$\mathcal{D}$, and
$\mathcal{R}_{\mathcal{D}}
=\{\mathcal{R}_i\}_{i=1}^{N_{\mathcal{D}}}$ is the complete routing
result, with
$\mathcal{R}_i\subseteq\mathcal{I}_{\mathcal{D}}$
denoting the issues assigned to $H_i^{+}$.


\subsection{Consultation}

The consultation stage performs problem-specific objective
design for each activated Department. Each activated Department $H_i^{+}$ maintains a catalog
$\mathcal{C}_i=\{m_{ij}\}_{j=1}^{J_i}$ of Loss Medicines. Each medicine
contains an implemented loss term, its intended effect, and its
applicability conditions. Guided by the Consultation Skill
$\mathcal{K}_{i}^{\mathrm{consult}}$, the Specialist Doctor selects
medicines that match the assigned issues $\mathcal{R}_i$:
\begin{equation}
\mathcal{M}_i
=
F_{\mathrm{consult}}
\left(
\mathcal{R}_i,
H_i^{+},
\mathcal{C}_i;
\mathcal{K}_{i}^{\mathrm{consult}}
\right),
\qquad
\mathcal{M}_i\subseteq\mathcal{C}_i.
\end{equation}
The selections from all activated Departments are collected as
$\mathcal{M}_{\mathcal{D}}
=\{\mathcal{M}_i\}_{i=1}^{N_{\mathcal{D}}}$
for subsequent objective construction.

\begin{table}[t]
\centering
\scriptsize
\setlength{\tabcolsep}{3.0pt}
\renewcommand{\arraystretch}{1.12}

\begin{tabularx}{\columnwidth}{
@{}
l
l
c
>{\raggedright\arraybackslash}X
@{}
}
\toprule
Dataset
& Domain
& \textit{V}/\textit{n}/\textit{d}/\textit{l}
& Views (dimensionality) \\
\midrule

SCENE
& Image
& 5/4400/634/33
& CH(64), CM(225), CORR(144), EDH(73), WT(128) \\

yeast
& Biology
& 2/2417/103/14
& GE(79), PP(24) \\

VOC07
& Image
& 3/3817/712/20
& DH(100), GIST(512), HH(100) \\

MIRFlickr
& Image
& 3/4053/712/38
& DH(100), GIST(512), HH(100) \\

mfeat
& Digits
& 6/2000/649/10
& FOU(76), FAC(216), KAR(64), PIX(240), ZER(47), MOR(6) \\

emotions
& Music
& 2/593/72/6
& RHY(8), TIM(64) \\

3Sources
& News
& 3/169/3000/6
& BBC(1000), Reuters(1000), Guardian(1000) \\

\bottomrule
\end{tabularx}
\caption{Detailed information of the datasets used in our experiments.
The statistics column reports the number of views, instances, features,
and labels, respectively.}
\label{tab:dataset_information}
\end{table}

\begin{table*}[t]
\centering
\setlength{\tabcolsep}{3.2pt}
\renewcommand{\arraystretch}{1.08}
\resizebox{\textwidth}{!}{%
\begin{tabular}{l|cccccccc}
\toprule
Dataset
& DHLI
& EF$^{2}$FS
& ENM
& GRAFS
& I$^{2}$VSLC
& MSFS
& LLM-Select
& FeatureHospital \\
\midrule

\multicolumn{9}{c}{\textit{AP} $\uparrow$} \\
\midrule

SCENE
& \res{0.7958}{0.0042}
& \res{0.7678}{0.0033}
& \res{0.7837}{0.0044}
& \res{0.8058}{0.0043}
& \second{0.8064}{0.0047}
& \res{0.7681}{0.0240}
& \best{0.8069}{0.0041}
& \res{0.8011}{0.0041} \\

yeast
& \res{0.6893}{0.0054}
& \res{0.6788}{0.0090}
& \res{0.6870}{0.0085}
& \res{0.6895}{0.0036}
& \res{0.6917}{0.0063}
& \res{0.6468}{0.0112}
& \best{0.7016}{0.0047}
& \second{0.6976}{0.0081} \\

VOC07
& \res{0.5885}{0.0081}
& \res{0.5067}{0.0116}
& \res{0.5941}{0.0070}
& \res{0.5858}{0.0052}
& \res{0.6003}{0.0032}
& \res{0.5082}{0.0103}
& \second{0.6070}{0.0049}
& \best{0.6073}{0.0042} \\

MIRFlickr
& \res{0.6621}{0.0060}
& \res{0.6706}{0.0058}
& \second{0.7025}{0.0048}
& \res{0.6741}{0.0038}
& \res{0.6634}{0.0059}
& \res{0.5606}{0.0465}
& \best{0.7027}{0.0039}
& \res{0.7011}{0.0048} \\

mfeat
& \res{0.6720}{0.0093}
& \res{0.4397}{0.1311}
& \res{0.8355}{0.0384}
& \res{0.8925}{0.0109}
& \second{0.9187}{0.0084}
& \res{0.7461}{0.0978}
& \res{0.8492}{0.0065}
& \best{0.9360}{0.0056} \\

emotions
& \res{0.5905}{0.0283}
& \res{0.6463}{0.0159}
& \best{0.6917}{0.0115}
& \res{0.6118}{0.0184}
& \res{0.6215}{0.0151}
& \res{0.5681}{0.0389}
& \res{0.6142}{0.0091}
& \second{0.6783}{0.0156} \\

3sources
& \res{0.3818}{0.0287}
& \res{0.3519}{0.0370}
& \res{0.3999}{0.0205}
& \res{0.3670}{0.0247}
& \res{0.3720}{0.0201}
& \second{0.4041}{0.0347}
& \res{0.3630}{0.0251}
& \best{0.4354}{0.0304} \\

\midrule
\multicolumn{9}{c}{\textit{AUC} $\uparrow$} \\
\midrule

SCENE
& \res{0.6637}{0.0041}
& \res{0.6119}{0.0052}
& \res{0.6476}{0.0071}
& \res{0.6954}{0.0048}
& \best{0.6975}{0.0049}
& \res{0.6236}{0.0423}
& \res{0.6965}{0.0051}
& \second{0.6970}{0.0050} \\

yeast
& \res{0.5911}{0.0090}
& \res{0.5780}{0.0077}
& \res{0.5975}{0.0083}
& \res{0.5866}{0.0056}
& \res{0.5921}{0.0063}
& \res{0.5335}{0.0077}
& \second{0.6111}{0.0035}
& \best{0.6196}{0.0083} \\

VOC07
& \res{0.6110}{0.0120}
& \res{0.5095}{0.0030}
& \res{0.6312}{0.0128}
& \res{0.6196}{0.0055}
& \res{0.6319}{0.0073}
& \res{0.4997}{0.0019}
& \second{0.6522}{0.0053}
& \best{0.6548}{0.0089} \\

MIRFlickr
& \res{0.5842}{0.0066}
& \res{0.6155}{0.0035}
& \second{0.6491}{0.0100}
& \res{0.6092}{0.0054}
& \res{0.5891}{0.0048}
& \res{0.5148}{0.0401}
& \best{0.6517}{0.0068}
& \res{0.6474}{0.0068} \\

mfeat
& \res{0.8954}{0.0063}
& \res{0.7076}{0.0945}
& \res{0.9568}{0.0190}
& \res{0.9714}{0.0046}
& \second{0.9807}{0.0032}
& \res{0.9060}{0.0591}
& \res{0.9581}{0.0024}
& \best{0.9871}{0.0024} \\

emotions
& \res{0.6160}{0.0263}
& \res{0.7271}{0.0136}
& \best{0.7582}{0.0120}
& \res{0.6580}{0.0133}
& \res{0.6628}{0.0222}
& \res{0.6167}{0.0412}
& \res{0.6572}{0.0091}
& \second{0.7326}{0.0169} \\

3sources
& \res{0.4898}{0.0236}
& \res{0.4993}{0.0040}
& \res{0.4994}{0.0209}
& \res{0.4983}{0.0255}
& \res{0.4893}{0.0312}
& \second{0.5010}{0.0031}
& \res{0.4891}{0.0242}
& \best{0.5109}{0.0177} \\

\bottomrule
\end{tabular}%
}
\caption{Comparison results in terms of AP and AUC
(mean $\pm$ standard deviation). Higher values indicate better
performance. The best and second-best results are highlighted in bold
and underlined, respectively.}
\label{tab:ap_auc_results}
\end{table*}

\subsection{Objective Construction}

After the Specialist Doctors prescribe the Loss Medicines for the activated Departments, the objective construction stage integrates these medicines into a unified feature selection objective. Similar to real-world medications, different Loss Medicines may exhibit conflicting effects and require appropriate dosage control. To address this issue, we design a Pharmacist Agent that reconciles the medicines prescribed by different Departments, removes redundant or conflicting terms, balances the weights of medicines, and integrates them into the final feature selection objective.

The Pharmacist coordinates the selected Loss Medicines according to the
Pharmacist Skill. Specifically, each medicine is assigned one of four
functional roles or a disabled status: (1) \emph{Backbone} loss provides the
main feature--label relevance signal; (2) \emph{Supporting} loss refines or
supplements the backbone signal; (3) \emph{Regularizer} loss captures
additional structures such as feature redundancy, label coverage,
feature quality, local structures, or view allocation; (4)
\emph{Guardrail} loss provides budget control or other defensive
constraints; and (5) \emph{Disabled} denotes medicines that are
redundant, conflicting, or unnecessary from the final objective. The
Pharmacist then balances the weights of the retained medicines and
integrates them into a compact final objective.

Formally, the objective construction process is defined as
\begin{equation}
\mathcal{L}_{\mathcal{D}}
=
F_{\mathrm{construct}}
\left(
\mathcal{M}_{\mathcal{D}},
P_{\mathcal{D}};
\mathcal{K}_{\mathrm{pharm}}
\right),
\end{equation}
where $\mathcal{M}_{\mathcal{D}}$ contains the Loss Medicines selected
by all activated Departments, $P_{\mathcal{D}}$ denotes the Statistical
Profile of dataset $\mathcal{D}$, and
$\mathcal{K}_{\mathrm{pharm}}$ denotes the Pharmacist Skill.

The resulting objective is expressed as
\begin{equation}
\begin{aligned}
\mathcal{L}_{\mathcal{D}}(z)
={}&
\lambda_{\mathrm{back}}
\mathcal{L}_{\mathrm{back}}(z)
+
\sum_{r=1}^{N_r}
\alpha_r
\mathcal{L}_{r}^{\mathrm{reg}}(z) \\
&+
\sum_{s=1}^{N_s}
\delta_s
\mathcal{L}_{s}^{\mathrm{sup}}(z)
+
\sum_{g=1}^{N_g}
\gamma_g
\mathcal{L}_{g}^{\mathrm{guard}}(z),
\end{aligned}
\end{equation}
where $\mathcal{L}_{\mathrm{back}}$ denotes the backbone loss, while
$\mathcal{L}_{r}^{\mathrm{reg}}$,
$\mathcal{L}_{s}^{\mathrm{sup}}$, and
$\mathcal{L}_{g}^{\mathrm{guard}}$ denote the retained regularization,
supporting, and guardrail losses, respectively. Here, $N_r$, $N_s$,
and $N_g$ are the numbers of retained losses in the corresponding
categories, where
$N_r,N_s,N_g\in\mathbb{Z}_{\geq 0}$. Therefore, the final objective
contains
$1+N_r+N_s+N_g$ loss terms. The
weights $\lambda_{\mathrm{back}}$,
$\{\alpha_r\}_{r=1}^{N_r}$,
$\{\delta_s\}_{s=1}^{N_s}$, and
$\{\gamma_g\}_{g=1}^{N_g}$ are assigned by the Pharmacist.

After the final objective is constructed, FeatureHospital optimizes it
on the training data to learn the importance of each feature. Specifically, the trainable parameter $\omega$ is instantiated as a logit vector $a\in\mathbb{R}^{d}$, from which a continuous feature-selection vector
is obtained:
\begin{equation}
z_j
=
\sigma
\left(
\frac{a_j}{\tau}
\right),
\qquad
j=1,\ldots,d,
\end{equation}
where $\sigma(\cdot)$ is the sigmoid function, $\tau$ is the
temperature parameter, and $z_j\in(0,1)$ denotes the selection strength
of the $j$-th feature.

The logit vector is optimized with respect to the constructed
dataset-specific objective:
\begin{equation}
a^{*}
=
\arg\min_{a\in\mathbb{R}^{d}}
\mathcal{L}_{\mathcal{D}}
\left(
z(a)
\right).
\end{equation}

After convergence, the optimized feature-selection vector is
$z^{*}=z(a^{*})$. Features are ranked according to their selection
strengths, and the top-$k$ subset is obtained as
\begin{equation}
\pi
=
\operatorname{argsort}
\left(
-z^{*}
\right),
\qquad
S_k
=
\left\{
\pi_1,\ldots,\pi_k
\right\},
\end{equation}
where $\pi=(\pi_1,\ldots,\pi_d)$ denotes the indices of all features
sorted in descending order of their selection strengths.
$S_k$ denotes the final selected feature subset containing the first
$k$ features in this ranking.

Finally, the selected feature subset is evaluated on the test set.
Detailed settings are provided in Appendix~D.

\subsection{General Skill Construction}

Throughout the workflow, each Agent is equipped with a dedicated Skill
that provides the knowledge and procedures required for its specialized
role. In our framework, \textbf{all Skills are
constructed from general-purpose data-analysis procedures or reusable
algorithmic components, rather than being designed or optimized for any
particular dataset.} For example,
the Skill library includes general mechanisms such as inverse-frequency
label weighting, correlation-based redundancy suppression, and
quadratic budget regularization. The underlying knowledge, formulas,
applicability conditions, and operating rules are defined in advance
and shared across datasets. Dataset-specific algorithms arise only when
the Agents select and combine suitable components from this fixed Skill
library during the workflow.

Complete Agent prompts, Skills, Department, medicine
catalogs, and validation rules are provided in Appendix~C.

\begin{figure}[t]
    \centering
    \includegraphics[width=0.9\columnwidth]{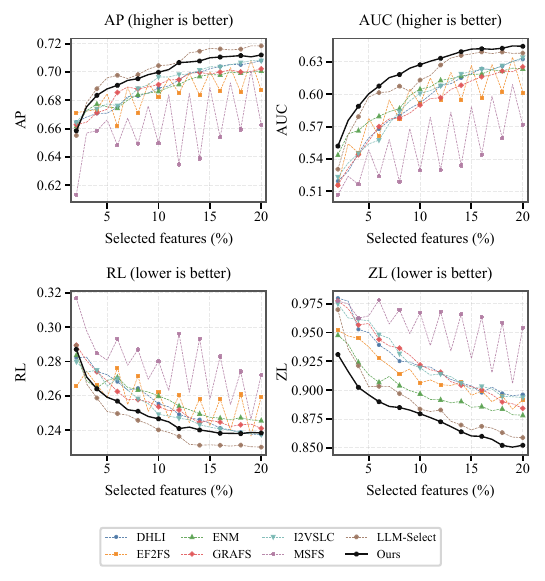}
    \caption{Performance comparison on yeast under different selected
    feature ratios ranging from 2\% to 20\%.}
    \label{fig:yeast_feature_ratio}
\end{figure}

\begin{table*}[t]
\centering
\setlength{\tabcolsep}{3.2pt}
\renewcommand{\arraystretch}{1.08}
\resizebox{\textwidth}{!}{%
\begin{tabular}{l|cccccccc}
\toprule
Dataset
& DHLI
& EF$^{2}$FS
& ENM
& GRAFS
& I$^{2}$VSLC
& MSFS
& LLM-Select
& FeatureHospital \\
\midrule

\multicolumn{9}{c}{\textit{RL} $\downarrow$} \\
\midrule

SCENE
& \res{0.1175}{0.0031}
& \res{0.1552}{0.0037}
& \res{0.1262}{0.0045}
& \best{0.1082}{0.0034}
& \second{0.1085}{0.0034}
& \res{0.1451}{0.0260}
& \res{0.1089}{0.0037}
& \res{0.1113}{0.0033} \\

yeast
& \res{0.2557}{0.0044}
& \res{0.2639}{0.0058}
& \res{0.2586}{0.0074}
& \res{0.2563}{0.0034}
& \res{0.2539}{0.0056}
& \res{0.2881}{0.0074}
& \best{0.2444}{0.0043}
& \second{0.2501}{0.0070} \\

VOC07
& \res{0.2796}{0.0087}
& \res{0.3917}{0.0212}
& \res{0.2738}{0.0086}
& \res{0.2798}{0.0051}
& \res{0.2655}{0.0035}
& \res{0.4007}{0.0166}
& \second{0.2567}{0.0048}
& \best{0.2544}{0.0059} \\

MIRFlickr
& \res{0.1944}{0.0047}
& \res{0.1956}{0.0046}
& \second{0.1689}{0.0035}
& \res{0.1874}{0.0030}
& \res{0.1939}{0.0045}
& \res{0.2567}{0.0254}
& \best{0.1681}{0.0029}
& \res{0.1716}{0.0040} \\

mfeat
& \res{0.1498}{0.0081}
& \res{0.5131}{0.1586}
& \res{0.0724}{0.0264}
& \res{0.0508}{0.0064}
& \second{0.0377}{0.0042}
& \res{0.1754}{0.1026}
& \res{0.0715}{0.0045}
& \best{0.0283}{0.0038} \\

emotions
& \res{0.4213}{0.0351}
& \res{0.3378}{0.0143}
& \best{0.2804}{0.0109}
& \res{0.3960}{0.0275}
& \res{0.3816}{0.0216}
& \res{0.4647}{0.0459}
& \res{0.3910}{0.0082}
& \second{0.2940}{0.0169} \\

3sources
& \res{0.5723}{0.0298}
& \res{0.6227}{0.0466}
& \res{0.5435}{0.0228}
& \res{0.5787}{0.0218}
& \res{0.5875}{0.0315}
& \second{0.5351}{0.0388}
& \res{0.5984}{0.0305}
& \best{0.5057}{0.0358} \\

\midrule
\multicolumn{9}{c}{\textit{ZL} $\downarrow$} \\
\midrule

SCENE
& \res{0.9488}{0.0034}
& \best{0.9255}{0.0048}
& \res{0.9626}{0.0023}
& \res{0.9372}{0.0039}
& \second{0.9341}{0.0039}
& \res{0.9540}{0.0134}
& \res{0.9361}{0.0031}
& \res{0.9399}{0.0033} \\

yeast
& \res{0.9230}{0.0050}
& \res{0.9143}{0.0089}
& \res{0.9010}{0.0085}
& \res{0.9231}{0.0082}
& \res{0.9248}{0.0045}
& \res{0.9665}{0.0065}
& \second{0.8919}{0.0098}
& \best{0.8790}{0.0122} \\

VOC07
& \res{0.9541}{0.0127}
& \res{0.9993}{0.0013}
& \res{0.9593}{0.0058}
& \res{0.9637}{0.0034}
& \res{0.9552}{0.0038}
& \res{0.9999}{0.0001}
& \best{0.9474}{0.0044}
& \second{0.9514}{0.0029} \\

MIRFlickr
& \res{0.9976}{0.0005}
& \res{0.9972}{0.0016}
& \second{0.9968}{0.0006}
& \res{0.9980}{0.0003}
& \res{0.9976}{0.0008}
& \res{0.9994}{0.0017}
& \res{0.9970}{0.0007}
& \best{0.9962}{0.0004} \\

mfeat
& \res{0.6687}{0.0090}
& \res{0.7438}{0.1224}
& \res{0.3474}{0.0726}
& \res{0.2206}{0.0191}
& \second{0.1807}{0.0209}
& \res{0.4263}{0.1140}
& \res{0.3072}{0.0098}
& \best{0.1364}{0.0099} \\

emotions
& \res{0.9179}{0.0258}
& \res{0.8588}{0.0161}
& \best{0.8147}{0.0184}
& \res{0.8904}{0.0236}
& \res{0.8901}{0.0136}
& \res{0.8849}{0.0323}
& \res{0.9000}{0.0098}
& \second{0.8452}{0.0197} \\

3sources
& \res{0.9929}{0.0146}
& \res{0.9953}{0.0099}
& \res{0.9957}{0.0063}
& \res{0.9988}{0.0014}
& \res{0.9988}{0.0025}
& \second{0.9816}{0.0293}
& \res{0.9982}{0.0025}
& \best{0.9522}{0.0221} \\

\bottomrule
\end{tabular}%
}
\caption{Comparison results in terms of RL and
ZL (mean $\pm$ standard deviation). Lower values indicate better
performance. The best and second-best results are highlighted in bold
and underlined, respectively. Rankings are determined using the
unrounded results.}
\label{tab:rl_01loss_results}
\end{table*}

\section{Experiments}

\subsection{Experiments Setup}

\paragraph{Datasets.}
We conduct experiments on seven benchmark datasets: SCENE~\citep{boutell2004learning}, yeast~\citep{elisseeff2001kernel}, VOC07~\citep{pascal-voc-2007}, MIRFlickr\citep{huiskes2008mir}, mfeat~\citep{multiple_features_72}, emotions~\citep{trohidis2008multi} and 3Sources~\citep{https://doi.org/10.1155/2021/5526479}. These datasets cover
five application domains, including handwritten digits, images, news,
biology, and music. Detailed dataset statistics are summarized in
Table~\ref{tab:dataset_information}.

\paragraph{Compared Methods.}
We compare FeatureHospital with six representative feature selection
methods, including DHLI~\citep{hao2024dhli}, EF$^{2}$FS~\citep{hao2025ef2fs}, ENM~\citep{gonzalez2020enm}, GRAFS~\citep{hao2024grafs}, I$^{2}$VSLC~\citep{hao2024i2vslc}, and
MSFS~\citep{zhang2020msfs}, as well as the LLM-based method LLM-Select~\citep{jeong2024llm}.

\paragraph{Evaluation Metrics.}
We adopt four widely used metrics for multi-label evaluation:
Average Precision (AP), Area Under the ROC Curve (AUC), Ranking Loss
(RL), and Zero--one Loss (ZL). Higher values indicate better performance for AP
and AUC, whereas lower values are preferred for RL and ZL. For each dataset, we randomly split the samples into 70\% for training
and 30\% for testing. Within each split, the results are averaged over feature
selection ratios from 2\% to 20\% at 2\% intervals. This process is repeated ten times, and the
results are reported as the mean $\pm$ standard deviation.

\paragraph{LLM Setup.}
Unless otherwise specified, all LLM-based agents in FeatureHospital use
GPT-5.5~\citep{openai2026gpt55} as the backbone model, with the temperature set to 0.1.

\begin{table*}[t]
\centering
\footnotesize
\setlength{\tabcolsep}{5.7pt}
\resizebox{0.83\textwidth}{!}{%
\begin{tabular}{llccccccc}
\toprule
Dataset
& Metric
& \shortstack{Full\\FeatureHospital}
& \shortstack{Backbone\\Only}
& \shortstack{Random\\Triage}
& \shortstack{Random\\Consultation}
& \shortstack{All Rule-Based\\Decisions}
& \shortstack{Single LLM\\Agent}
& \shortstack{Without\\Pharmacist} \\
\midrule

\multirow{4}{*}{SCENE}
& AP $\uparrow$
& \bestmean{0.8011}
& 0.7892
& 0.7956
& 0.7893
& 0.7994
& 0.7963
& \secondmean{0.8003} \\

& AUC $\uparrow$
& \bestmean{0.6970}
& 0.6784
& 0.6891
& 0.6768
& 0.6954
& 0.6905
& \secondmean{0.6962} \\

& RL $\downarrow$
& \bestmean{0.1113}
& 0.1187
& 0.1145
& 0.1187
& 0.1122
& 0.1142
& \secondmean{0.1118} \\

& ZL $\downarrow$
& \bestmean{0.9399}
& 0.9543
& 0.9468
& 0.9538
& \secondmean{0.9400}
& 0.9438
& 0.9418 \\

\midrule

\multirow{4}{*}{mfeat}
& AP $\uparrow$
& \bestmean{0.9360}
& 0.9071
& 0.8700
& 0.9217
& 0.8867
& 0.9217
& \secondmean{0.9235} \\

& AUC $\uparrow$
& \bestmean{0.9871}
& 0.9801
& 0.9617
& 0.9831
& 0.9716
& 0.9830
& \secondmean{0.9836} \\

& RL $\downarrow$
& \bestmean{0.0283}
& 0.0412
& 0.0660
& 0.0351
& 0.0534
& 0.0347
& \secondmean{0.0342} \\

& ZL $\downarrow$
& \bestmean{0.1364}
& 0.1898
& 0.2625
& \secondmean{0.1595}
& 0.2319
& 0.1661
& 0.1599 \\

\midrule

\multirow{4}{*}{3Sources}
& AP $\uparrow$
& \secondmean{0.4354}
& 0.4135
& 0.4175
& 0.4010
& 0.3920
& 0.4179
& \bestmean{0.4364} \\

& AUC $\uparrow$
& \bestmean{0.5109}
& 0.4932
& 0.5009
& 0.4975
& 0.5042
& 0.4919
& \secondmean{0.5079} \\

& RL $\downarrow$
& \bestmean{0.5057}
& 0.5224
& 0.5270
& 0.5383
& 0.5420
& 0.5276
& \secondmean{0.5128} \\

& ZL $\downarrow$
& \bestmean{0.9522}
& 0.9776
& 0.9731
& 0.9785
& 0.9927
& 0.9690
& \secondmean{0.9555} \\

\bottomrule
\end{tabular}
}
\caption{
Ablation results in terms of AP, AUC, RL, and ZL. The best and second-best results are highlighted in bold and underlined, respectively.
}
\label{tab:ablation_results}
\end{table*}

\subsection{Main Results}

\subsubsection{Comparison Results.}

Tables~\ref{tab:ap_auc_results} and
\ref{tab:rl_01loss_results} compare FeatureHospital with seven methods on seven datasets. Two main observations can be observed. (1) FeatureHospital achieves strong
overall performance across the four evaluation metrics. It obtains the
best or second-best results on multiple datasets, while remaining close to
the best-performing method in the other cases. In particular, on
mfeat and 3sources, FeatureHospital consistently outperforms all compared methods
across AP, AUC, RL, and ZL. These results demonstrate that FeatureHospital can automatically design effective feature selection algorithms for different datasets.  (2) FeatureHospital demonstrates more consistent performance across different datasets. For example, LLM-Select performs strongly on several
datasets, including SCENE, yeast, VOC07, and MIRFlickr, but exhibits
substantial performance degradation on mfeat, emotions, and 3Sources.
In contrast, FeatureHospital remains competitive across all seven
datasets without suffering a performance collapse on any
particular dataset. In addition, Figure~\ref{fig:yeast_feature_ratio} presents the results on yeast with selected feature ratios ranging from 2\% to
20\%. The corresponding results on the remaining datasets are provided
in Appendix~B.

Overall, these results demonstrate that FeatureHospital achieves strong performance while maintaining consistent effectiveness across dataset changes.

\begin{figure}[t]
    \centering
    \includegraphics[width=0.86\linewidth]{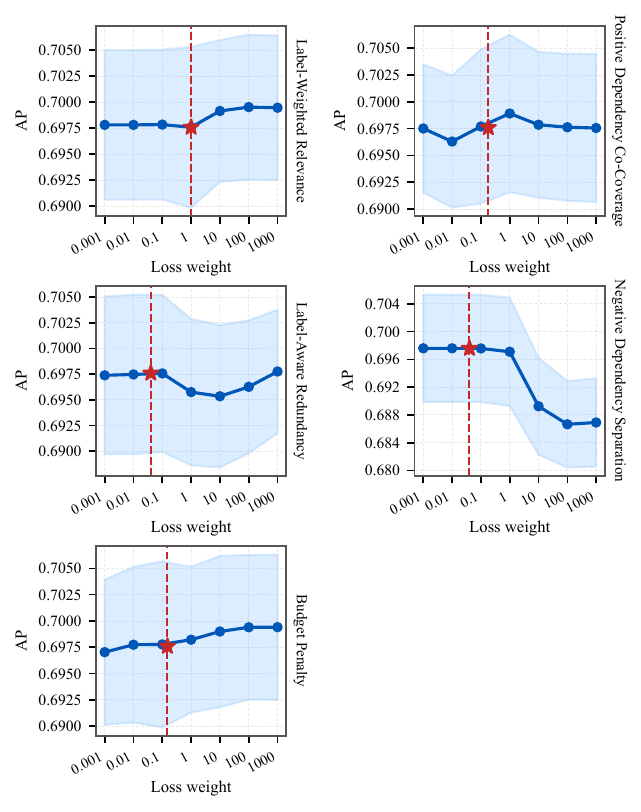}
    \caption{Parameter sensitivity of the objective constructed for yeast.}
    \label{fig:yeast_parameter_sensitivity}
\end{figure}

\subsubsection{Parameter Analysis}

We perform a parameter sensitivity analysis on yeast,
using AP as the evaluation metric. The objective constructed for yeast
contains Label-Weighted Relevance as the backbone loss, Positive
Dependency Co-Coverage as the supporting loss, Label-Aware Pairwise
Redundancy as the regularization loss, and Negative Dependency
Separation and Budget Penalty as the guardrail losses. For each term,
its weight is varied over
$\{0.001,0.01,0.1,1,10,100,1000\}$ while all other weights remain
fixed. As shown in Figure~\ref{fig:yeast_parameter_sensitivity},
several terms remain effective across broad weight ranges, whereas
Negative Dependency Separation is more sensitive to excessively large
values. The weights selected by FeatureHospital lie within stable
regions.

\subsection{Ablation Study}

Table~\ref{tab:ablation_results} evaluates FeatureHospital from four
perspectives. Results on the remaining datasets are provided in Appendix~B.

\textbf{(1) Algorithm effectiveness.} \textbf{Backbone
Only} removes all non-backbone terms from the final objective. Its
performance degradation confirms that the additional loss terms provide
effective complementary constraints.

\textbf{(2) Effectiveness of individual pipeline stages.}
\textbf{Random Triage} activates three random Departments and averages
up to ten configurations; \textbf{Random Consultation} selects two
random Loss Medicines per activated Department and averages up to six
configurations; and \textbf{Without Pharmacist} directly combines the
default medicines and weights recommended by the Specialist Doctors.
When fewer configurations are available, all possible configurations
are used. The full method generally outperforms these variants,
demonstrating the effectiveness of each stage. The smaller gap without the Pharmacist suggests that the
Specialist Doctors already provide reasonable initial configurations,
which are further coordinated and refined by the Pharmacist.

\textbf{(3) Effectiveness of LLM-based decisions.} \textbf{All
Rule-Based Decisions} replaces all LLM decisions with predefined rules extracted from the corresponding Skills.
Its lower performance, particularly on mfeat and 3Sources, demonstrates
the importance of involving LLMs in decisions
throughout the algorithm-design process. Additional LLM
replacement experiments are reported in Appendix~B, while Appendix~E
presents case studies with manual inspection of agent outputs to further
examine the validity of LLM decisions.

\textbf{(4) Effectiveness of the overall multi-agent pipeline.}
\textbf{Single LLM Agent} collapses the entire staged decision process
into a single LLM call. Given the Statistical Profile of the target
dataset and the complete catalog of candidate Loss Medicines, one agent
directly selects and combines the medicines into a final
dataset-specific objective. Its consistently lower performance supports
the effectiveness of the role specialization and staged collaboration
in FeatureHospital.

\section{Conclusion}
We presented FeatureHospital, a Skill-driven multi-agent framework for
automated MVML feature selection algorithm customization. By diagnosing
dataset issues, selecting problem-specific Loss Medicines, and
reconciling them into a compact objective, FeatureHospital constructs
different feature selection algorithms for different datasets.
Experiments on seven datasets demonstrate competitive and consistent
performance, while the ablation study validate the contributions of
the objective components and multi-agent workflow.
\newpage
\bibliography{aaai2027}

\newpage
\appendix
\setcounter{secnumdepth}{2}
\setcounter{tocdepth}{2}
\section{Threats \& Discussion}
\label{app:discussion}

\subsection{Threats}

\paragraph{Dependence on a Static, Human-Curated Skill Library.}
FeatureHospital currently relies on a collection of predefined Skills
constructed by human experts. These Skills specify the domain knowledge,
applicability conditions, operating procedures, and candidate Loss Medicines
used by the agents throughout diagnosis, triage, consultation, and objective
construction. Although this design provides a controllable and interpretable
foundation for automated algorithm customization, the capability of
FeatureHospital is inevitably bounded by the coverage and quality of the
existing Skill library. In particular, when a dataset exhibits an issue that
is not adequately represented by the available Skills, the framework may fail
to identify an appropriate treatment or construct a sufficiently expressive
objective. An important future direction is therefore to investigate automatic
Skill acquisition and refinement. For example, new Skills may be extracted
from research papers, existing implementations, and accumulated experimental
results, followed by systematic validation, deduplication, and integration into
the existing Skill library. Such mechanisms would reduce the dependence on
manual Skill engineering and continuously expand the range of problems that
FeatureHospital can address.

\paragraph{Lack of Experience-Driven Self-Evolution.}
FeatureHospital is designed to construct a dataset-specific feature selection
algorithm for each input dataset, but it does not yet accumulate experience
across repeated runs. The diagnostic decisions, Department routing results,
selected Loss Medicines, constructed objectives, and their empirical outcomes
are not persistently retained to improve subsequent algorithm-design processes.
Consequently, even when the framework encounters datasets with similar
characteristics, it cannot directly reuse previously successful prescriptions
or learn systematically from unsuccessful ones. Future work will explore an
experience-driven self-evolution mechanism that records dataset profiles,
agent decisions, constructed objectives, and evaluation feedback in a
persistent experience memory. When processing a new dataset, relevant prior
cases could be retrieved to support diagnosis and prescription, while
performance feedback could be used to refine decision rules, Skill
applicability conditions, and medicine configurations. This would enable
FeatureHospital to evolve from a framework that only applies predefined
expertise into one that progressively improves through accumulated algorithm
design experience.

\subsection{Discussion: Automatic Feature Selection}
Automated machine learning (AutoML) aims to automate the construction
of machine learning solutions, thereby reducing the expert effort
required in algorithm design, selection, and configuration
\cite{he2021automl}. When applied to feature selection, this idea leads
to automatic feature selection, which aims to automatically construct
a feature selection algorithm tailored to the characteristics of the
target dataset.

Existing studies have primarily approached this problem through
algorithm recommendation and pipeline search. Early work characterizes
a target dataset using meta-features, identifies similar historical
datasets, and ranks a predefined collection of feature selection
algorithms according to their previous performance
\cite{wang2014feature}. Subsequent meta-learning frameworks learn the
relationship between dataset characteristics and the relative
performance of candidate feature selection algorithms, thereby
recommending a suitable algorithm or configuration for a new dataset
\cite{parmezan2017metalearning,parmezan2021automatic}. More recent work
further considers combinations of predefined components. For example,
FSPL jointly recommends a filter method and an embedded method as a
feature selection pipeline rather than selecting the two stages
independently \cite{lazebnik2023fspl}. Although these approaches differ
in their dataset representations and recommendation mechanisms, they
generally operate over a predefined search space whose candidates are
complete feature selection algorithms, algorithm configurations, or
pipelines assembled from predefined stages.

FeatureHospital differs from these approaches in the granularity and
organization of algorithm construction. Rather than directly selecting
a complete algorithm or pipeline from predefined candidates,
FeatureHospital first decomposes the feature selection requirements of
the target dataset into individual issues, such as label imbalance,
feature redundancy, and view-quality imbalance. It then addresses these
issues one by one by selecting a corresponding optimization component
for each identified problem. Finally, the selected components are
reconciled to remove overlaps and conflicts, their contributions are
balanced, and they are progressively assembled into a unified
dataset-specific objective. Existing approaches can therefore be viewed
as performing coarse-grained recommendation or search at the algorithm
or pipeline level, whereas FeatureHospital performs fine-grained,
problem-driven construction at the objective-component level. The former
asks which predefined feature selection solution should be applied to
the target dataset, while the latter asks which problems are present,
how each problem should be modeled, and how the resulting treatments
should be integrated into a coherent feature selection algorithm.

\section{Additional Experiments}
\label{app:experiments}




\begin{table*}[t]
\centering
\small
\rmfamily
\setlength{\tabcolsep}{3pt}
\renewcommand{\arraystretch}{1.15}

\begin{tabular*}{\textwidth}{
  @{\extracolsep{\fill}}
  lcccccc
  @{}
}
\toprule

\headleft{Dataset}
&
\headone{GPT-5.5$^{\dagger}$}
&
\headtwo{Qwen3.5}{27B}
&
\headtwo{GPT-5.4}{mini}
&
\headtwo{DeepSeek}{V4-Pro}
&
\headtwo{Claude}{Sonnet 5}
&
\headtwo{Gemini 3.1}{Pro Preview}
\\

\midrule
\multicolumn{7}{c}{\textit{AP} $\uparrow$} \\
\midrule

SCENE
& \second{0.8011}{0.0041}
& \best{0.8039}{0.0040}
& \val{0.7872}{0.0037}
& \val{0.7875}{0.0041}
& \val{0.7952}{0.0034}
& \val{0.7951}{0.0041}
\\

yeast
& \second{0.6976}{0.0081}
& \val{0.6908}{0.0065}
& \val{0.6915}{0.0069}
& \best{0.6989}{0.0082}
& \val{0.6934}{0.0079}
& \val{0.6910}{0.0074}
\\

VOC07
& \val{0.6073}{0.0042}
& \val{0.6064}{0.0042}
& \second{0.6074}{0.0042}
& \val{0.5964}{0.0050}
& \best{0.6078}{0.0051}
& \val{0.6068}{0.0043}
\\

MIRFlickr
& \val{0.7011}{0.0048}
& \val{0.7008}{0.0046}
& \val{0.7008}{0.0045}
& \val{0.6991}{0.0045}
& \second{0.7019}{0.0048}
& \best{0.7019}{0.0044}
\\

mfeat
& \second{0.9360}{0.0056}
& \val{0.9235}{0.0080}
& \val{0.9328}{0.0070}
& \val{0.9213}{0.0098}
& \val{0.9242}{0.0093}
& \best{0.9534}{0.0054}
\\

emotions
& \val{0.6783}{0.0156}
& \val{0.6781}{0.0166}
& \val{0.6783}{0.0148}
& \best{0.6807}{0.0180}
& \second{0.6797}{0.0168}
& \val{0.6769}{0.0148}
\\

3Sources
& \best{0.4354}{0.0304}
& \val{0.4106}{0.0350}
& \val{0.4342}{0.0263}
& \val{0.4089}{0.0256}
& \second{0.4343}{0.0365}
& \val{0.4308}{0.0316}
\\

\specialrule{\lightrulewidth}{0pt}{1.5pt}

\avglabel
& \avgbest{0.6938}
& \avgval{0.6877}
& \avgval{0.6903}
& \avgval{0.6847}
& \avgval{0.6909}
& \avgsecond{0.6937}
\\

\specialrule{\lightrulewidth}{1.5pt}{1.5pt}

\multicolumn{7}{c}{\textit{AUC} $\uparrow$}
\\

\specialrule{\lightrulewidth}{1.5pt}{0pt}

SCENE
& \second{0.6970}{0.0050}
& \best{0.6993}{0.0054}
& \val{0.6735}{0.0061}
& \val{0.6742}{0.0062}
& \val{0.6873}{0.0065}
& \val{0.6880}{0.0066}
\\

yeast
& \second{0.6196}{0.0083}
& \val{0.6067}{0.0048}
& \val{0.6100}{0.0047}
& \best{0.6216}{0.0068}
& \val{0.6118}{0.0086}
& \val{0.6087}{0.0051}
\\

VOC07
& \second{0.6548}{0.0089}
& \val{0.6530}{0.0084}
& \best{0.6551}{0.0087}
& \val{0.6312}{0.0096}
& \val{0.6528}{0.0065}
& \val{0.6494}{0.0096}
\\

MIRFlickr
& \val{0.6474}{0.0068}
& \val{0.6464}{0.0065}
& \val{0.6472}{0.0065}
& \val{0.6432}{0.0067}
& \best{0.6489}{0.0067}
& \second{0.6487}{0.0062}
\\

mfeat
& \val{0.9871}{0.0024}
& \val{0.9828}{0.0041}
& \second{0.9872}{0.0024}
& \val{0.9818}{0.0045}
& \val{0.9834}{0.0032}
& \best{0.9917}{0.0022}
\\

emotions
& \val{0.7326}{0.0169}
& \val{0.7333}{0.0173}
& \val{0.7333}{0.0165}
& \best{0.7363}{0.0179}
& \second{0.7337}{0.0172}
& \val{0.7324}{0.0160}
\\

3Sources
& \best{0.5109}{0.0177}
& \second{0.5102}{0.0228}
& \val{0.5055}{0.0222}
& \val{0.5069}{0.0307}
& \val{0.5081}{0.0217}
& \val{0.5040}{0.0188}
\\

\specialrule{\lightrulewidth}{0pt}{1.5pt}

\avglabel
& \avgbest{0.6928}
& \avgsecond{0.6902}
& \avgval{0.6874}
& \avgval{0.6850}
& \avgval{0.6894}
& \avgval{0.6890}
\\

\bottomrule
\end{tabular*}

\caption{Robustness of FeatureHospital across different LLM
backbones in terms of AP and AUC (mean $\pm$ standard deviation).
GPT-5.5$^{\dagger}$ denotes the reference backbone used in the main
experiments. Higher values indicate better performance. The best and
second-best results are highlighted in bold and underlined,
respectively. Rankings are determined using the unrounded results.}
\label{tab:llm-backbone-ap-auc}
\end{table*}

\begin{table*}[t]
\centering
\small
\rmfamily
\setlength{\tabcolsep}{3pt}
\renewcommand{\arraystretch}{1.15}

\begin{tabular*}{\textwidth}{
  @{\extracolsep{\fill}}
  lcccccc
  @{}
}
\toprule

\headleft{Dataset}
&
\headone{GPT-5.5$^{\dagger}$}
&
\headtwo{Qwen3.5}{27B}
&
\headtwo{GPT-5.4}{mini}
&
\headtwo{DeepSeek}{V4-Pro}
&
\headtwo{Claude}{Sonnet 5}
&
\headtwo{Gemini 3.1}{Pro Preview}
\\

\midrule
\multicolumn{7}{c}{\textit{RL} $\downarrow$} \\
\midrule

SCENE
& \second{0.1113}{0.0033}
& \best{0.1100}{0.0028}
& \val{0.1200}{0.0033}
& \val{0.1196}{0.0036}
& \val{0.1155}{0.0033}
& \val{0.1150}{0.0037}
\\

yeast
& \second{0.2501}{0.0070}
& \val{0.2591}{0.0060}
& \val{0.2574}{0.0062}
& \best{0.2490}{0.0063}
& \val{0.2554}{0.0076}
& \val{0.2586}{0.0061}
\\

VOC07
& \best{0.2544}{0.0059}
& \val{0.2552}{0.0058}
& \second{0.2544}{0.0059}
& \val{0.2672}{0.0067}
& \val{0.2547}{0.0055}
& \val{0.2565}{0.0068}
\\

MIRFlickr
& \val{0.1716}{0.0040}
& \val{0.1720}{0.0042}
& \val{0.1716}{0.0040}
& \val{0.1731}{0.0041}
& \second{0.1709}{0.0040}
& \best{0.1700}{0.0036}
\\

mfeat
& \second{0.0283}{0.0038}
& \val{0.0353}{0.0064}
& \val{0.0283}{0.0046}
& \val{0.0367}{0.0070}
& \val{0.0340}{0.0054}
& \best{0.0214}{0.0036}
\\

emotions
& \val{0.2940}{0.0169}
& \val{0.2929}{0.0183}
& \val{0.2938}{0.0162}
& \best{0.2902}{0.0196}
& \second{0.2921}{0.0181}
& \val{0.2941}{0.0167}
\\

3Sources
& \best{0.5057}{0.0358}
& \val{0.5232}{0.0365}
& \val{0.5153}{0.0281}
& \val{0.5257}{0.0391}
& \val{0.5079}{0.0423}
& \second{0.5078}{0.0373}
\\

\specialrule{\lightrulewidth}{0pt}{1.5pt}

\avglabel
& \avgbest{0.2308}
& \avgval{0.2354}
& \avgval{0.2344}
& \avgval{0.2374}
& \avgval{0.2329}
& \avgsecond{0.2319}
\\

\specialrule{\lightrulewidth}{1.5pt}{1.5pt}

\multicolumn{7}{c}{\textit{ZL} $\downarrow$}
\\

\specialrule{\lightrulewidth}{1.5pt}{0pt}

SCENE
& \second{0.9399}{0.0033}
& \best{0.9346}{0.0029}
& \val{0.9562}{0.0018}
& \val{0.9566}{0.0016}
& \val{0.9445}{0.0034}
& \val{0.9442}{0.0026}
\\

yeast
& \val{0.8790}{0.0122}
& \val{0.8778}{0.0095}
& \val{0.8794}{0.0108}
& \best{0.8769}{0.0115}
& \val{0.8797}{0.0110}
& \second{0.8776}{0.0097}
\\

VOC07
& \val{0.9514}{0.0029}
& \second{0.9505}{0.0030}
& \val{0.9516}{0.0030}
& \val{0.9605}{0.0026}
& \best{0.9505}{0.0037}
& \val{0.9512}{0.0030}
\\

MIRFlickr
& \second{0.9962}{0.0004}
& \val{0.9962}{0.0007}
& \val{0.9965}{0.0008}
& \val{0.9968}{0.0006}
& \best{0.9961}{0.0009}
& \val{0.9962}{0.0010}
\\

mfeat
& \second{0.1364}{0.0099}
& \val{0.1555}{0.0108}
& \val{0.1444}{0.0120}
& \val{0.1608}{0.0133}
& \val{0.1604}{0.0151}
& \best{0.1027}{0.0108}
\\

emotions
& \val{0.8452}{0.0197}
& \val{0.8467}{0.0206}
& \best{0.8448}{0.0193}
& \second{0.8448}{0.0214}
& \val{0.8460}{0.0208}
& \val{0.8460}{0.0206}
\\

3Sources
& \second{0.9522}{0.0221}
& \val{0.9863}{0.0088}
& \best{0.9512}{0.0354}
& \val{0.9808}{0.0220}
& \val{0.9604}{0.0267}
& \val{0.9594}{0.0310}
\\

\specialrule{\lightrulewidth}{0pt}{1.5pt}

\avglabel
& \avgsecond{0.8143}
& \avgval{0.8211}
& \avgval{0.8177}
& \avgval{0.8253}
& \avgval{0.8197}
& \avgbest{0.8110}
\\

\bottomrule
\end{tabular*}

\caption{Robustness of FeatureHospital across different LLM
backbones in terms of RL and ZL (mean $\pm$ standard deviation).
GPT-5.5$^{\dagger}$ denotes the reference backbone used in the main
experiments. Lower values indicate better performance. The best and
second-best results are highlighted in bold and underlined,
respectively. Rankings are determined using the unrounded results.}
\label{tab:llm-backbone-rl-zl}
\end{table*}

\subsection{Backbone LLM Replacement}
\label{app:backbone_replacement}

To examine whether FeatureHospital depends on a particular backbone
LLM, we replace the GPT-5.5~\cite{openai2026gpt55} backbone used in the main experiments with
models covering different capability levels and accessibility
settings. These include two proprietary high-capability models, Claude
Sonnet 5~\citep{claude} and Gemini 3.1 Pro Preview~\citep{gemini}; one open-source flagship model,
DeepSeek V4-Pro~\citep{deepseekv4}; one proprietary compact model, GPT-5.4 mini~\citep{openai2026gpt54mini}; and one
compact open-source model, Qwen3.5-27B~\citep{qwen3.5}. The Skills, agent prompts,
available tools, and all subsequent training and evaluation settings
are kept unchanged. Tables~\ref{tab:llm-backbone-ap-auc} and
\ref{tab:llm-backbone-rl-zl} report the resulting performance.

The overall performance of FeatureHospital is relatively insensitive
to the choice of backbone LLM. All six backbones achieve competitive
results across the seven datasets, and replacing GPT-5.5 does not cause
a substantial performance degradation or failure on any evaluation
metric. Across the six backbones, the differences between the best and
worst average results are only 0.0091 for AP, 0.0078 for AUC, 0.0066
for RL, and 0.0143 for ZL. This consistency suggests that the structured
workflow provided by the Skills and specialized agents constrains the
algorithm-design process sufficiently well for different LLMs to
construct effective feature selection objectives.

Relatively lightweight backbones also retain competitive performance.
For example, GPT-5.4-mini achieves average AP, AUC, RL, and ZL values
of 0.6903, 0.6874, 0.2344, and 0.8177, respectively, which remain
close to those obtained with GPT-5.5. Qwen3.5-27B similarly achieves
competitive average results and performs best on SCENE under several
metrics. These results indicate that FeatureHospital can potentially
use smaller or lower-cost backbone models to reduce LLM inference
expenses while preserving most of its feature selection effectiveness,
providing flexibility for deployments with different computational and
budget constraints.

GPT-5.5 nevertheless provides the most stable and balanced overall
performance. It achieves the best average results for AP, AUC, and RL,
with values of 0.6938, 0.6928, and 0.2308, respectively, and obtains
the second-best average ZL of 0.8143. Other backbones occasionally
perform better on individual datasets, such as DeepSeek-V4-Pro on
yeast and Gemini-3.1-Pro-Preview on mfeat, but none consistently
dominates across datasets and metrics. Therefore, GPT-5.5 remains a
suitable default backbone for the main experiments, while the results
with alternative LLMs demonstrate that the effectiveness of
FeatureHospital is not tightly coupled to this particular model.

\paragraph{Cost of a Single Pipeline Execution.}
To evaluate the practical cost of using FeatureHospital, we record the
average number of LLM calls and token consumption required by one
complete pipeline execution. Based on the standard online API prices
available at the time of evaluation, we further estimate the
corresponding monetary cost, as reported in
Table~\ref{tab:single-run-cost}. The estimation uses regular real-time
inference prices without applying prompt caching, batch processing, or
other promotional discounts.

\begin{table}[h]
\centering
\scriptsize
\rmfamily
\setlength{\tabcolsep}{2.2pt}
\renewcommand{\arraystretch}{1.15}

\begin{tabular*}{\columnwidth}{
  @{\extracolsep{\fill}}
  lrrrrr
  @{}
}
\toprule
Model
& Calls
& \makecell{Input\\(K)}
& \makecell{Output\\(K)}
& \makecell{Total\\(K)}
& \makecell{Cost\\(USD)}
\\
\midrule

Qwen3.5-27B
& 9.0
& 132.7
& 21.0
& 153.8
& \$0.090
\\

DeepSeek-V4-Pro
& 9.1
& 136.7
& 20.6
& 157.4
& \$0.079
\\

GPT-5.4 mini
& 10.3
& 159.8
& 49.0
& 208.8
& \$0.340
\\

\makecell[l]{Gemini 3.1 Pro\\Preview}
& 8.9
& 166.0
& 36.6
& 202.6
& \$0.771
\\

Claude Sonnet 5
& 10.3
& 243.9
& 82.1
& 326.1
& \$1.309
\\

\bottomrule
\end{tabular*}

\caption{Average LLM usage and estimated API cost of one complete
FeatureHospital pipeline execution. K denotes one thousand tokens.
Costs are calculated using the standard real-time API prices available
at the time of evaluation without prompt-caching or batch-processing
discounts.}
\label{tab:single-run-cost}
\end{table}

Although a complete pipeline execution involves multiple interactions
among the specialized agents, it only needs to be performed once to
construct a complete dataset-specific feature selection objective and
its corresponding algorithm configuration. The resulting algorithm can
then be optimized and applied to the target dataset without repeating
the LLM-based algorithm-design process. As shown in
Table~\ref{tab:single-run-cost}, the average cost of one complete
execution ranges from approximately \$0.08 to \$1.31. In particular,
Qwen3.5-27B and GPT-5.4 mini require only approximately \$0.09 and
\$0.34, respectively, while retaining competitive feature selection
performance. Considering that a single execution replaces the manual
process of dataset diagnosis, objective design, and component
coordination, this one-time cost is modest and practically acceptable
for constructing a customized feature selection algorithm.

\begin{figure*}[t]
    \centering
    \includegraphics[width=\textwidth]{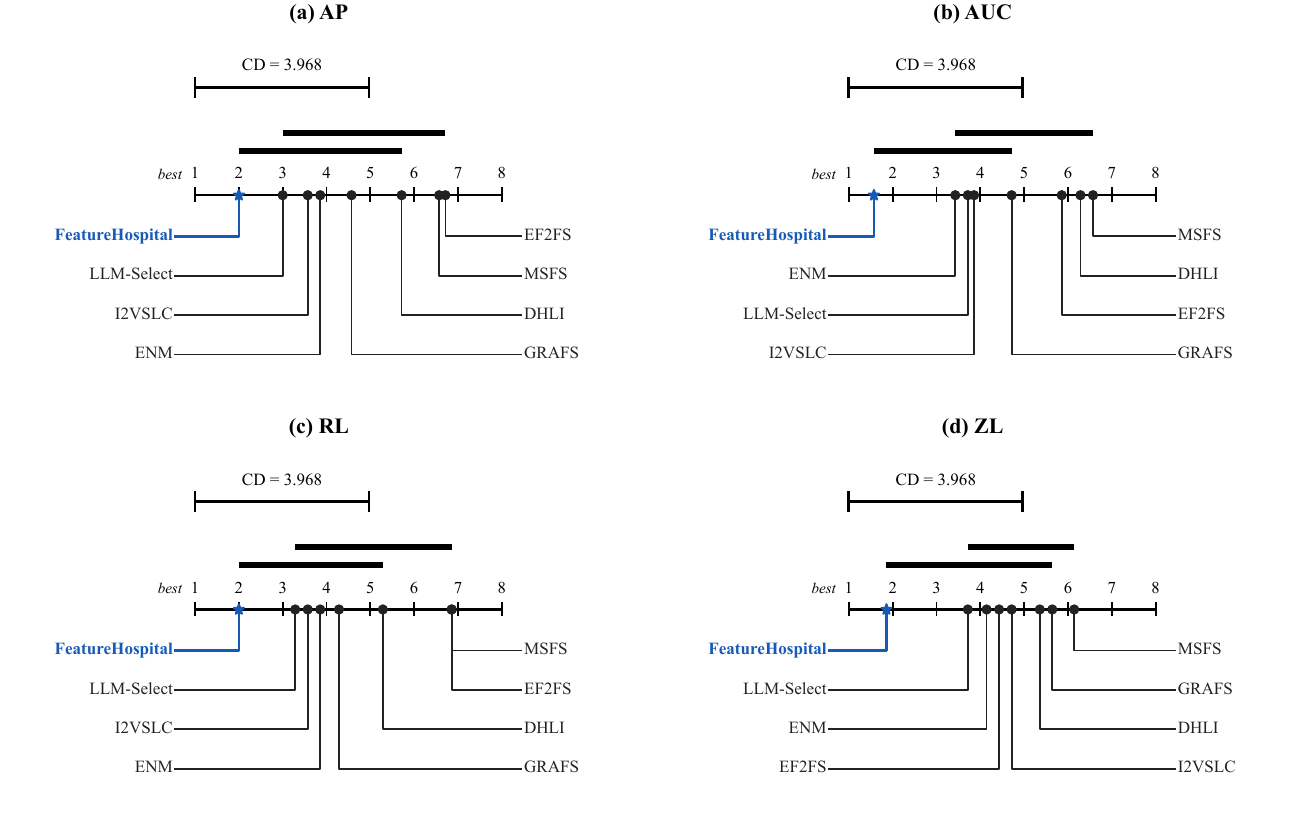}
    \caption{Critical difference diagrams for AP, AUC, RL, and ZL over
    the seven datasets. Lower average ranks indicate better overall
    performance. Methods connected by the same thick horizontal line
    are not significantly different according to the Nemenyi post-hoc
    test at $\alpha=0.05$ ($\mathrm{CD}=3.968$).}
    \label{fig:cd_diagrams}
\end{figure*}

\subsection{Statistical Significance Analysis}
\label{app:significance}

To further examine whether the performance differences among the
compared methods are statistically significant across datasets, we
follow the standard procedure proposed by Demšar
\citep{demvsar2006statistical}. Specifically, we first conduct the
Friedman test for each evaluation metric and then apply the Nemenyi
post-hoc test for pairwise comparisons. For each dataset, the compared
methods are ranked according to their performance, where rank 1 is
assigned to the best-performing method. AP and AUC are ranked in
descending order, whereas RL and ZL are ranked in ascending order. The
average rank of each method is then computed over the seven datasets.

For the eight compared methods and seven datasets, the critical
difference at the significance level $\alpha=0.05$ is calculated as
\begin{equation}
\mathrm{CD}
=
q_{\alpha}
\sqrt{\frac{k(k+1)}{6N}}
=
3.968,
\end{equation}
where $k=8$ is the number of methods and $N=7$ is the number of
datasets. In the critical difference diagrams, methods connected by the
same thick horizontal line do not exhibit statistically significant
differences under the Nemenyi test.

As shown in Figure~\ref{fig:cd_diagrams}, FeatureHospital achieves the
best average rank under all four evaluation metrics. Its average ranks
are 2.000, 1.571, 2.000, and 1.857 for AP, AUC, RL, and ZL,
respectively. The Friedman tests reject the null hypothesis that all
methods have equivalent performance for AP ($p=0.0012$), AUC
($p=0.0014$), RL ($p=0.0010$), and ZL ($p=0.0407$), indicating that
statistically significant performance differences exist among the
compared methods.

The subsequent Nemenyi tests provide more detailed pairwise
comparisons. FeatureHospital significantly outperforms EF2FS and MSFS
in terms of AP and RL. For AUC, it significantly outperforms DHLI,
EF2FS, and MSFS, while for ZL, it significantly outperforms MSFS.
The differences between FeatureHospital and several strong baselines,
such as LLM-Select, I2VSLC, ENM, and GRAFS, do not exceed the critical
difference on all metrics and are therefore not statistically
significant under the Nemenyi test.

Overall, the statistical analysis shows that FeatureHospital
consistently obtains the best overall ranking across different datasets
and evaluation metrics. Although the relatively small number of
datasets results in a large critical difference and limits the
statistical power of some pairwise comparisons, FeatureHospital
maintains consistently competitive performance without exhibiting a
substantial degradation on any particular metric.

\subsection{Performance across Feature Selection Ratios}
\label{app:performance_curves}

\begin{figure*}[t]
    \centering
    \includegraphics[width=0.49\textwidth]
    {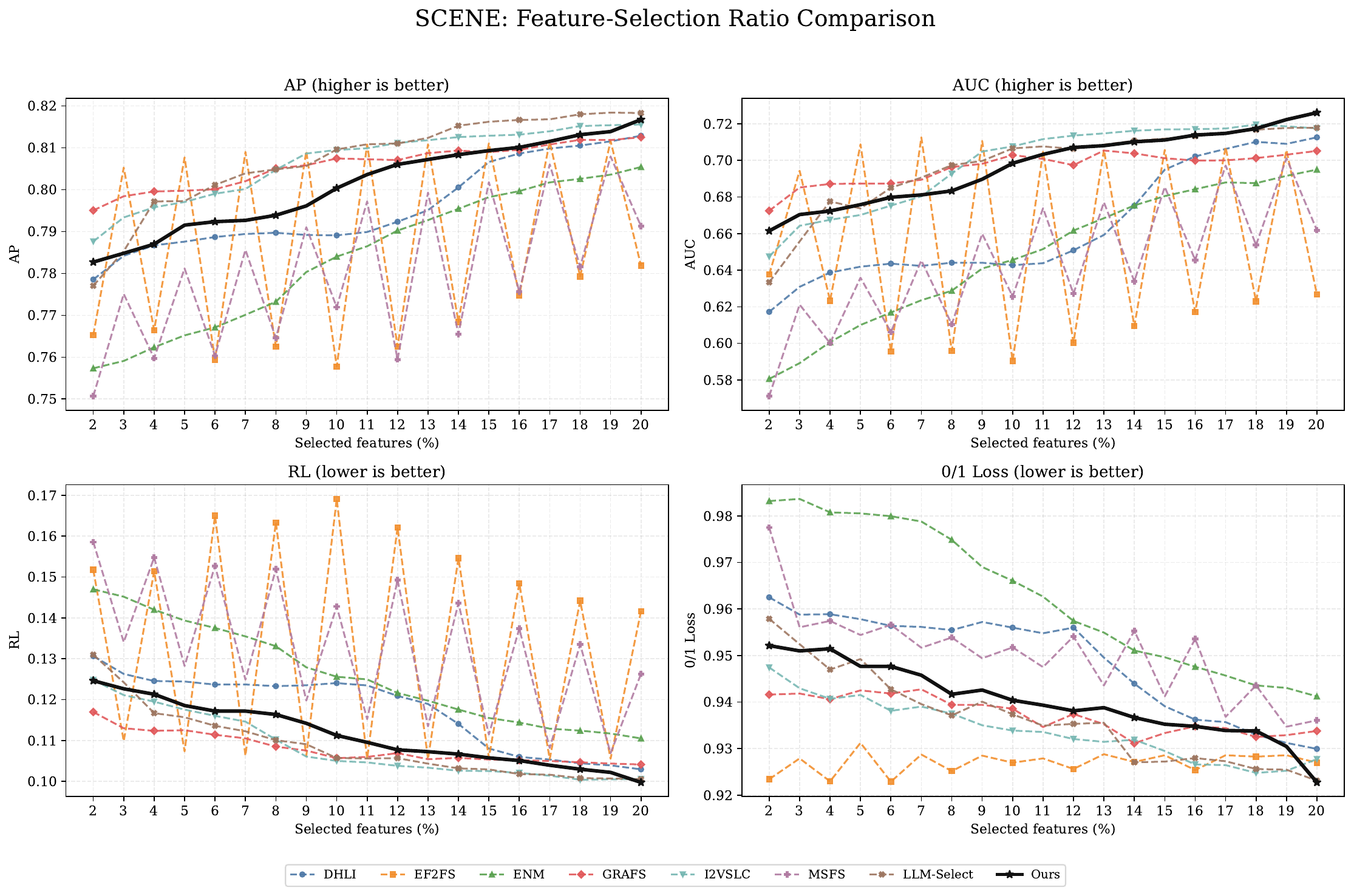}
    \hfill
    \includegraphics[width=0.49\textwidth]
    {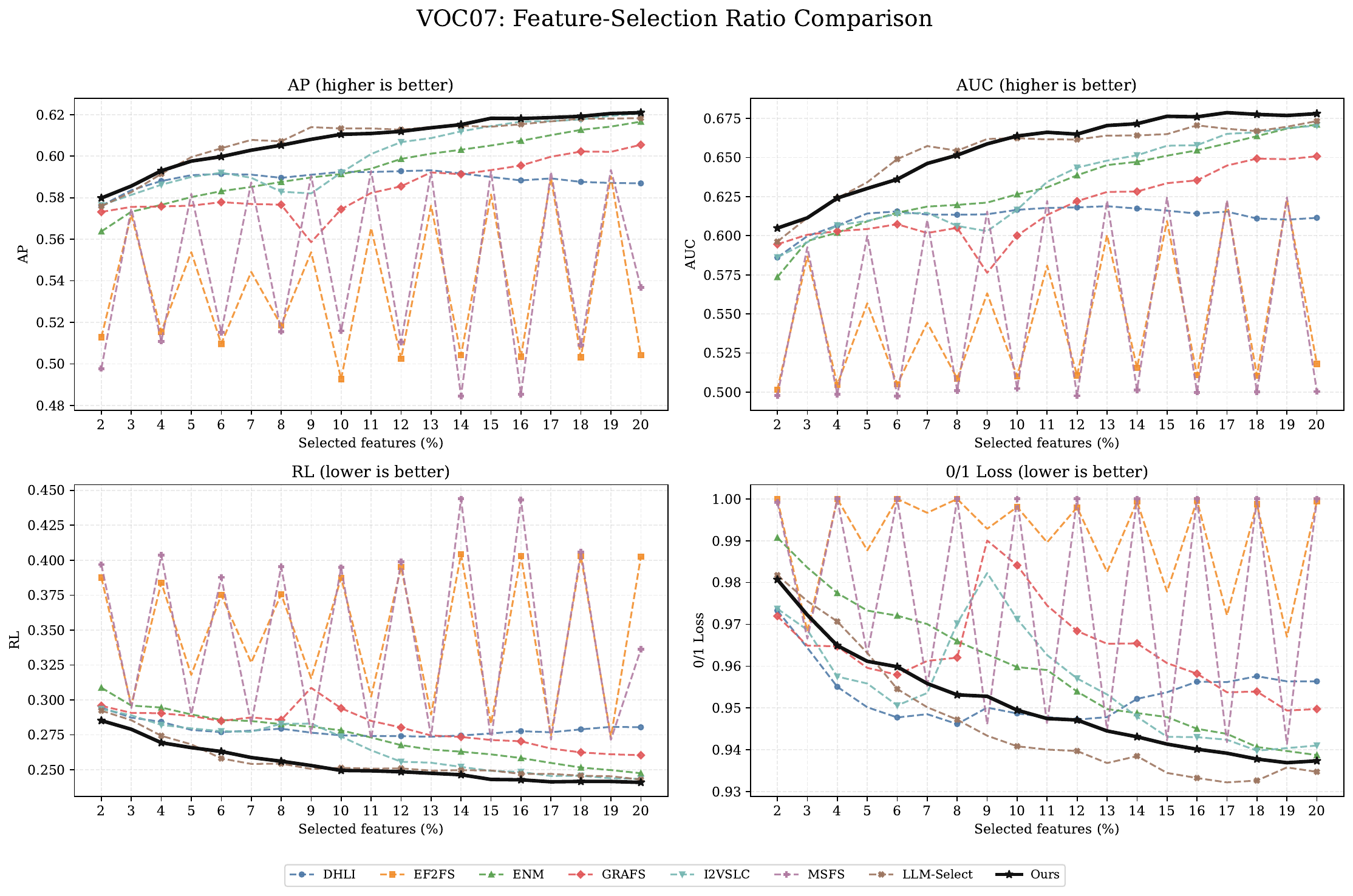}
    \caption{Performance under different feature selection ratios on
    SCENE and VOC07. Higher values are preferred for AP and AUC,
    whereas lower values are preferred for RL and ZL.}
    \label{fig:ratio-scene-voc07}
\end{figure*}

\begin{figure*}[t]
    \centering
    \includegraphics[width=0.49\textwidth]
    {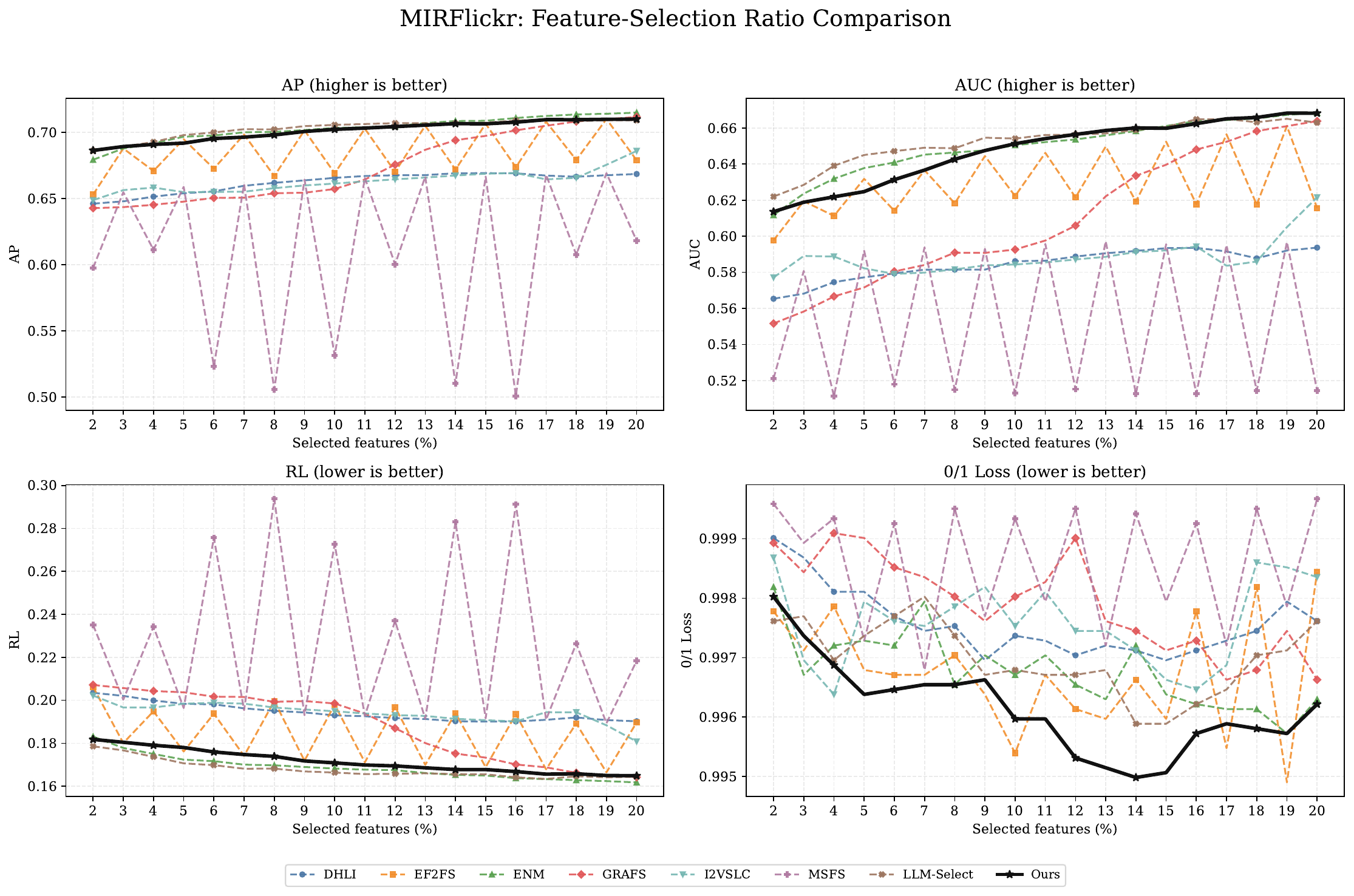}
    \hfill
    \includegraphics[width=0.49\textwidth]
    {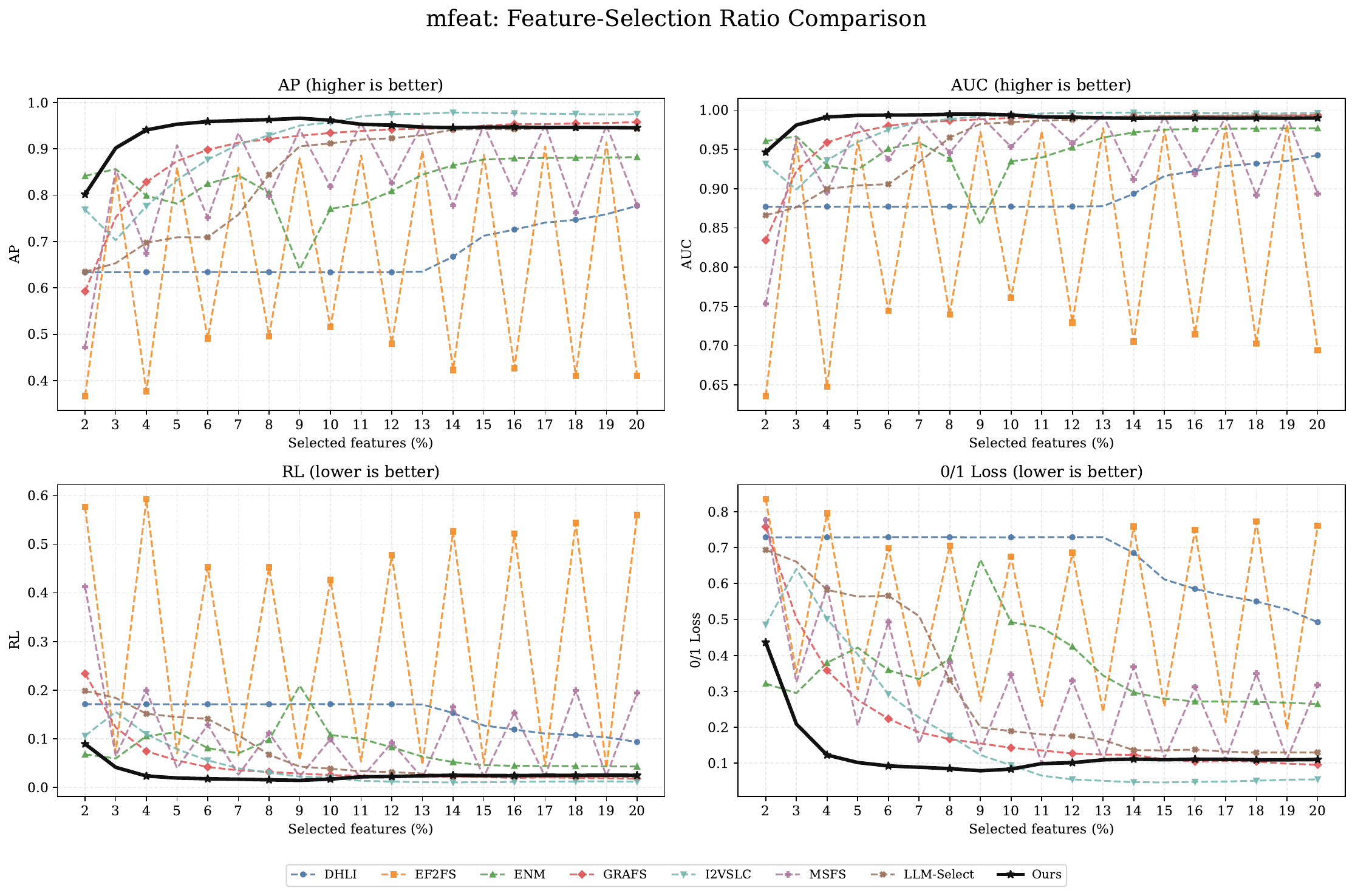}
    \caption{Performance under different feature selection ratios on
    MIRFlickr and mfeat. Higher values are preferred for AP and AUC,
    whereas lower values are preferred for RL and ZL.}
    \label{fig:ratio-mirflickr-mfeat}
\end{figure*}

\begin{figure*}[t]
    \centering
    \includegraphics[width=0.49\textwidth]
    {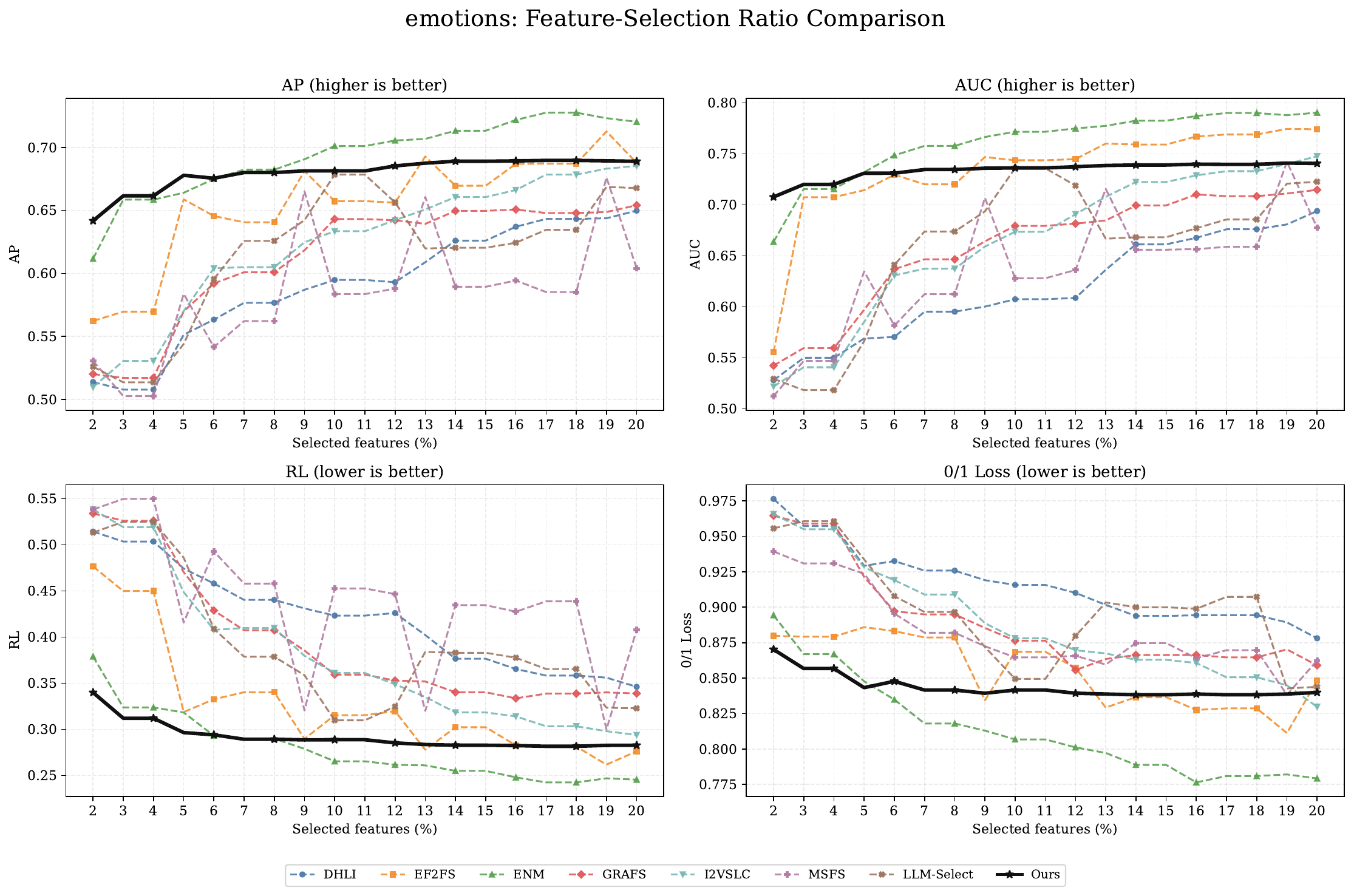}
    \hfill
    \includegraphics[width=0.49\textwidth]
    {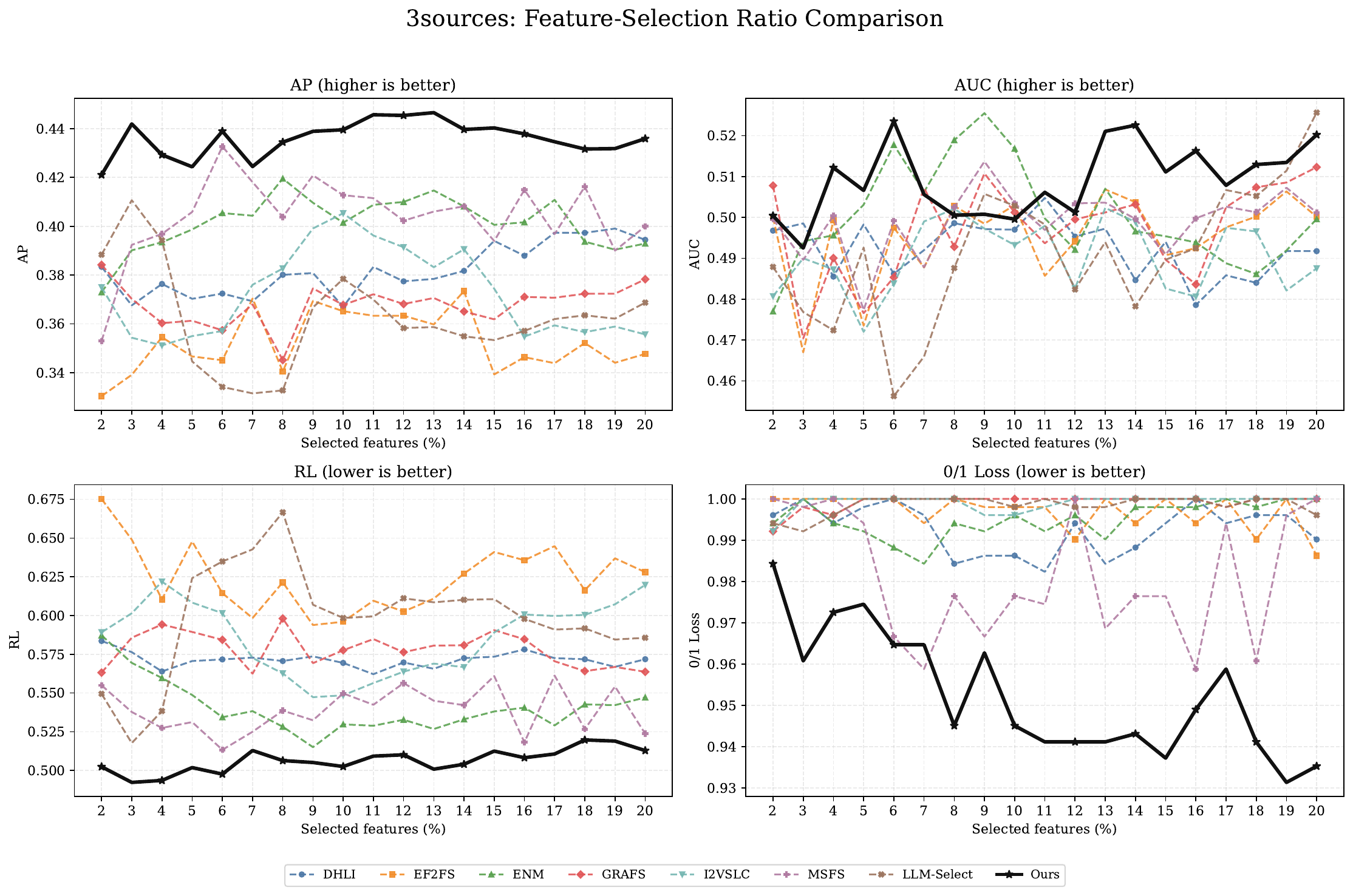}
    \caption{Performance under different feature selection ratios on
    emotions and 3Sources. Higher values are preferred for AP and AUC,
    whereas lower values are preferred for RL and ZL.}
    \label{fig:ratio-emotions-3sources}
\end{figure*}

The main text reports the performance curves on yeast as a
representative example. To provide a more complete comparison,
Figures~\ref{fig:ratio-scene-voc07}--\ref{fig:ratio-emotions-3sources}
present the results on the remaining six datasets under feature
selection ratios ranging from 2\% to 20\%. For each ratio, all methods
are evaluated using AP, AUC, RL, and ZL.

Across different feature selection ratios, FeatureHospital generally
maintains leading or competitive performance rather than performing
well only under a particular feature budget. Its advantage is especially
clear on mfeat and 3Sources, while it also remains close to the
best-performing methods on the other datasets. Although individual
baselines may achieve better results at isolated ratios or on particular
metrics, no consistent degradation of FeatureHospital is observed as
the feature budget changes.

The performance of FeatureHospital generally improves or remains stable
as more features are selected. More importantly, it already achieves
strong results at relatively small feature selection ratios, indicating
that the constructed objectives assign high importance to informative
features and can produce effective compact subsets. After a moderate
number of features has been selected, the performance often approaches
a stable region, suggesting that the most useful information has already
been retained.

FeatureHospital also exhibits relatively smooth performance curves
across consecutive feature selection ratios. In contrast, several
competing methods show substantial fluctuations when the number of
selected features changes slightly. This indicates that the feature
rankings produced by FeatureHospital are less sensitive to the exact
feature budget and provide more consistent subsets across different
selection ratios. Together with the yeast results reported in the main
text, these observations demonstrate that the effectiveness of
FeatureHospital is robust across both datasets and feature selection
ratios.

\begin{table*}[t]
\centering
\small
\rmfamily
\setlength{\tabcolsep}{2pt}
\renewcommand{\arraystretch}{1.10}

\begin{tabular*}{\textwidth}{
  @{\extracolsep{\fill}}
  lccccccc
  @{}
}
\toprule

\headleft{Dataset}
&
\headtwo{Full}{FeatureHospital}
&
\headtwo{Backbone}{Only}
&
\headtwo{Random}{Triage}
&
\headtwo{Random}{Consultation}
&
\headtwo{Without}{Pharmacist}
&
\headtwo{All Rule-Based}{Decisions}
&
\headtwo{Single LLM}{Agent}
\\

\midrule
\multicolumn{8}{c}{\textit{AP} $\uparrow$} \\
\midrule

\rowstrut SCENE
& \best{0.8011}{0.0041}
& \val{0.7892}{0.0039}
& \val{0.7956}{0.0062}
& \val{0.7893}{0.0042}
& \second{0.8003}{0.0042}
& \val{0.7994}{0.0044}
& \val{0.7963}{0.0043}
\\

\rowstrut yeast
& \val{0.6976}{0.0081}
& \best{0.6995}{0.0073}
& \val{0.6949}{0.0097}
& \val{0.6910}{0.0084}
& \val{0.6975}{0.0078}
& \val{0.6973}{0.0080}
& \second{0.6977}{0.0083}
\\

\rowstrut VOC07
& \best{0.6073}{0.0042}
& \val{0.6051}{0.0039}
& \val{0.6019}{0.0094}
& \val{0.6053}{0.0048}
& \second{0.6073}{0.0045}
& \val{0.5936}{0.0045}
& \val{0.6014}{0.0042}
\\

\rowstrut MIRFlickr
& \val{0.7011}{0.0048}
& \val{0.7003}{0.0044}
& \second{0.7015}{0.0045}
& \val{0.6887}{0.0167}
& \val{0.7011}{0.0047}
& \val{0.6985}{0.0043}
& \best{0.7017}{0.0041}
\\

\rowstrut mfeat
& \best{0.9360}{0.0056}
& \val{0.9071}{0.0056}
& \val{0.8700}{0.0905}
& \val{0.9217}{0.0213}
& \second{0.9235}{0.0077}
& \val{0.8867}{0.0056}
& \val{0.9217}{0.0074}
\\

\rowstrut emotions
& \val{0.6783}{0.0156}
& \val{0.6783}{0.0156}
& \second{0.6793}{0.0155}
& \val{0.6787}{0.0142}
& \val{0.6783}{0.0139}
& \best{0.6796}{0.0137}
& \val{0.6772}{0.0117}
\\

\rowstrut 3Sources
& \second{0.4354}{0.0304}
& \val{0.4135}{0.0277}
& \val{0.4175}{0.0320}
& \val{0.4010}{0.0307}
& \best{0.4364}{0.0294}
& \val{0.3920}{0.0286}
& \val{0.4179}{0.0344}
\\

\specialrule{\lightrulewidth}{1.0pt}{1.0pt}

\rowstrut\textbf{Average}
& \avgbest{0.6938}
& \avgval{0.6847}
& \avgval{0.6801}
& \avgval{0.6823}
& \avgsecond{0.6921}
& \avgval{0.6781}
& \avgval{0.6877}
\\

\addlinespace[2pt]
\midrule
\multicolumn{8}{c}{\textit{AUC} $\uparrow$} \\
\midrule

\rowstrut SCENE
& \best{0.6970}{0.0050}
& \val{0.6784}{0.0058}
& \val{0.6891}{0.0105}
& \val{0.6768}{0.0069}
& \second{0.6962}{0.0053}
& \val{0.6954}{0.0071}
& \val{0.6905}{0.0075}
\\

\rowstrut yeast
& \second{0.6196}{0.0083}
& \best{0.6217}{0.0058}
& \val{0.6149}{0.0136}
& \val{0.6075}{0.0099}
& \val{0.6196}{0.0076}
& \val{0.6192}{0.0064}
& \val{0.6180}{0.0087}
\\

\rowstrut VOC07
& \second{0.6548}{0.0089}
& \val{0.6511}{0.0093}
& \val{0.6443}{0.0187}
& \val{0.6503}{0.0120}
& \best{0.6561}{0.0095}
& \val{0.6226}{0.0077}
& \val{0.6400}{0.0105}
\\

\rowstrut MIRFlickr
& \second{0.6474}{0.0068}
& \val{0.6456}{0.0066}
& \val{0.6473}{0.0061}
& \val{0.6281}{0.0248}
& \val{0.6471}{0.0067}
& \val{0.6423}{0.0065}
& \best{0.6476}{0.0071}
\\

\rowstrut mfeat
& \best{0.9871}{0.0024}
& \val{0.9801}{0.0022}
& \val{0.9617}{0.0403}
& \val{0.9831}{0.0083}
& \second{0.9836}{0.0030}
& \val{0.9716}{0.0020}
& \val{0.9830}{0.0030}
\\

\rowstrut emotions
& \val{0.7326}{0.0169}
& \val{0.7326}{0.0169}
& \val{0.7338}{0.0164}
& \val{0.7334}{0.0155}
& \best{0.7350}{0.0164}
& \second{0.7347}{0.0174}
& \val{0.7344}{0.0161}
\\

\rowstrut 3Sources
& \best{0.5109}{0.0177}
& \val{0.4932}{0.0175}
& \val{0.5009}{0.0225}
& \val{0.4975}{0.0204}
& \second{0.5079}{0.0250}
& \val{0.5042}{0.0310}
& \val{0.4919}{0.0232}
\\

\specialrule{\lightrulewidth}{1.0pt}{1.0pt}

\rowstrut\textbf{Average}
& \avgbest{0.6928}
& \avgval{0.6861}
& \avgval{0.6846}
& \avgval{0.6824}
& \avgsecond{0.6922}
& \avgval{0.6843}
& \avgval{0.6865}
\\

\bottomrule
\end{tabular*}

\caption{Ablation results of FeatureHospital in terms of AP and AUC
(mean $\pm$ standard deviation). Higher values indicate better
performance. The Average row reports the arithmetic mean across the
seven datasets. The best and second-best results are highlighted in
bold and underlined, respectively. Rankings are determined using the
unrounded results.}
\label{tab:ablation-ap-auc}
\end{table*}

\begin{table*}[t]
\centering
\small
\rmfamily
\setlength{\tabcolsep}{2pt}
\renewcommand{\arraystretch}{1.10}

\begin{tabular*}{\textwidth}{
  @{\extracolsep{\fill}}
  lccccccc
  @{}
}
\toprule

\headleft{Dataset}
&
\headtwo{Full}{FeatureHospital}
&
\headtwo{Backbone}{Only}
&
\headtwo{Random}{Triage}
&
\headtwo{Random}{Consultation}
&
\headtwo{Without}{Pharmacist}
&
\headtwo{All Rule-Based}{Decisions}
&
\headtwo{Single LLM}{Agent}
\\

\midrule
\multicolumn{8}{c}{\textit{RL} $\downarrow$} \\
\midrule

\rowstrut SCENE
& \best{0.1113}{0.0033}
& \val{0.1187}{0.0034}
& \val{0.1145}{0.0046}
& \val{0.1187}{0.0037}
& \second{0.1118}{0.0036}
& \val{0.1122}{0.0036}
& \val{0.1142}{0.0039}
\\

\rowstrut yeast
& \val{0.2501}{0.0070}
& \best{0.2492}{0.0056}
& \val{0.2537}{0.0100}
& \val{0.2573}{0.0087}
& \second{0.2500}{0.0065}
& \val{0.2515}{0.0068}
& \val{0.2507}{0.0069}
\\

\rowstrut VOC07
& \second{0.2544}{0.0059}
& \val{0.2567}{0.0056}
& \val{0.2609}{0.0116}
& \val{0.2572}{0.0071}
& \best{0.2542}{0.0062}
& \val{0.2741}{0.0053}
& \val{0.2628}{0.0059}
\\

\rowstrut MIRFlickr
& \second{0.1716}{0.0040}
& \val{0.1724}{0.0041}
& \val{0.1717}{0.0039}
& \val{0.1785}{0.0093}
& \val{0.1716}{0.0040}
& \val{0.1737}{0.0040}
& \best{0.1704}{0.0040}
\\

\rowstrut mfeat
& \best{0.0283}{0.0038}
& \val{0.0412}{0.0036}
& \val{0.0660}{0.0559}
& \val{0.0351}{0.0117}
& \second{0.0342}{0.0047}
& \val{0.0535}{0.0031}
& \val{0.0347}{0.0047}
\\

\rowstrut emotions
& \val{0.2940}{0.0169}
& \val{0.2939}{0.0169}
& \best{0.2925}{0.0169}
& \val{0.2934}{0.0153}
& \val{0.2943}{0.0143}
& \second{0.2934}{0.0145}
& \val{0.2948}{0.0136}
\\

\rowstrut 3Sources
& \best{0.5057}{0.0358}
& \val{0.5224}{0.0298}
& \val{0.5270}{0.0389}
& \val{0.5383}{0.0302}
& \second{0.5128}{0.0351}
& \val{0.5420}{0.0292}
& \val{0.5276}{0.0331}
\\

\specialrule{\lightrulewidth}{1.0pt}{1.0pt}

\rowstrut\textbf{Average}
& \avgbest{0.2308}
& \avgval{0.2363}
& \avgval{0.2409}
& \avgval{0.2398}
& \avgsecond{0.2327}
& \avgval{0.2429}
& \avgval{0.2364}
\\

\addlinespace[2pt]
\midrule
\multicolumn{8}{c}{\textit{ZL} $\downarrow$} \\
\midrule

\rowstrut SCENE
& \best{0.9399}{0.0033}
& \val{0.9543}{0.0015}
& \val{0.9468}{0.0064}
& \val{0.9538}{0.0039}
& \val{0.9418}{0.0036}
& \second{0.9400}{0.0027}
& \val{0.9438}{0.0028}
\\

\rowstrut yeast
& \val{0.8790}{0.0122}
& \best{0.8752}{0.0114}
& \second{0.8765}{0.0109}
& \val{0.8808}{0.0109}
& \val{0.8798}{0.0118}
& \val{0.8781}{0.0105}
& \val{0.8781}{0.0119}
\\

\rowstrut VOC07
& \best{0.9514}{0.0029}
& \val{0.9538}{0.0034}
& \val{0.9571}{0.0087}
& \val{0.9539}{0.0046}
& \second{0.9516}{0.0032}
& \val{0.9657}{0.0034}
& \val{0.9548}{0.0025}
\\

\rowstrut MIRFlickr
& \best{0.9962}{0.0004}
& \val{0.9964}{0.0005}
& \second{0.9962}{0.0008}
& \val{0.9971}{0.0011}
& \val{0.9965}{0.0004}
& \val{0.9969}{0.0008}
& \val{0.9968}{0.0008}
\\

\rowstrut mfeat
& \best{0.1364}{0.0099}
& \val{0.1898}{0.0114}
& \val{0.2625}{0.1785}
& \second{0.1595}{0.0415}
& \val{0.1599}{0.0124}
& \val{0.2320}{0.0110}
& \val{0.1661}{0.0109}
\\

\rowstrut emotions
& \val{0.8452}{0.0197}
& \val{0.8452}{0.0197}
& \val{0.8451}{0.0193}
& \val{0.8445}{0.0189}
& \val{0.8435}{0.0180}
& \best{0.8421}{0.0183}
& \second{0.8425}{0.0191}
\\

\rowstrut 3Sources
& \best{0.9522}{0.0221}
& \val{0.9776}{0.0236}
& \val{0.9731}{0.0258}
& \val{0.9785}{0.0260}
& \second{0.9555}{0.0269}
& \val{0.9927}{0.0085}
& \val{0.9690}{0.0312}
\\

\specialrule{\lightrulewidth}{1.0pt}{1.0pt}

\rowstrut\textbf{Average}
& \avgbest{0.8143}
& \avgval{0.8275}
& \avgval{0.8368}
& \avgval{0.8240}
& \avgsecond{0.8184}
& \avgval{0.8354}
& \avgval{0.8216}
\\

\bottomrule
\end{tabular*}

\caption{Ablation results of FeatureHospital in terms of RL and ZL
(mean $\pm$ standard deviation). Lower values indicate better
performance. The Average row reports the arithmetic mean across the
seven datasets. The best and second-best results are highlighted in
bold and underlined, respectively. Rankings are determined using the
unrounded results.}
\label{tab:ablation-rl-zl}
\end{table*}

\subsection{Complete Ablation Results}
\label{app:ablation_results}

The main text reports ablation results on SCENE, mfeat, and 3Sources
as representative cases. Tables~\ref{tab:ablation-ap-auc} and
\ref{tab:ablation-rl-zl} provide the complete results on all seven
datasets. Although individual variants may occasionally achieve the
best result on a particular dataset or metric, the complete
FeatureHospital consistently obtains the best average performance
across all four metrics. It achieves average AP, AUC, RL, and ZL
values of 0.6938, 0.6928, 0.2308, and 0.8143, respectively,
demonstrating that the full design provides the most reliable overall
performance across datasets.

The \emph{Backbone Only} variant removes all supporting,
regularization, and guardrail terms from the constructed objective.
Its average performance decreases to 0.6847 AP and 0.6861 AUC, while
RL and ZL increase to 0.2363 and 0.8275. The degradation is
particularly evident on mfeat and 3Sources, showing that the additional
Loss Medicines selected for dataset-specific issues provide important
constraints beyond the basic feature--label relevance signal.
Although the backbone objective alone can remain effective on some
datasets, it cannot consistently address the diverse combinations of
feature selection problems encountered across datasets.

Randomizing either the triage or consultation stage also leads to
clear performance degradation. \emph{Random Triage} may activate
Departments that are unrelated to the diagnosed issues, whereas
\emph{Random Consultation} may select Loss Medicines that do not match
the assigned problems. Both variants obtain worse average results than
the complete method under all four metrics. Their degradation and, in
some cases, substantially larger standard deviations on mfeat and
MIRFlickr indicate that appropriate issue routing and problem-specific
medicine selection are important not only for effectiveness but also
for the stability of the constructed objectives.

Replacing all LLM-based decisions with predefined rules produces the
weakest or nearly weakest average performance on most metrics. The
\emph{All Rule-Based Decisions} variant achieves only 0.6781 AP and
0.6843 AUC, together with 0.2429 RL and 0.8354 ZL. This result suggests
that fixed rules extracted from the Skills cannot fully capture the
context-dependent interactions among dataset characteristics,
diagnostic evidence, and candidate Loss Medicines. LLM-based reasoning
is therefore useful for adapting the general knowledge encoded in the
Skills to the specific conditions of each dataset.

The \emph{Single LLM Agent} variant also performs consistently worse
than the complete framework on average, despite occasionally obtaining
strong results on individual datasets such as MIRFlickr. Collapsing
diagnosis, routing, consultation, and objective reconciliation into one
decision removes the explicit intermediate structure and role-specific
context provided by the multi-agent workflow. Its lower average
performance supports the use of specialized agents and staged
collaboration rather than a single monolithic LLM call.

Among all ablations, \emph{Without Pharmacist} is the strongest
variant, achieving average results close to those of the complete
framework. This indicates that the Specialist Doctors already provide
generally reasonable initial medicines and weights. Nevertheless, the
complete method remains better on average for every metric, showing
that the Pharmacist provides a consistent additional benefit by
removing overlaps, resolving conflicts, and balancing the contributions
of independently prescribed objective terms. Overall, the complete
ablation results confirm that the effectiveness of FeatureHospital
arises from the combined contributions of dataset-specific objective
components, structured triage and consultation, LLM-based decisions,
multi-agent specialization, and final objective reconciliation.

\section{Skills \& Prompt}
\label{app:prompt}

This appendix specifies the prompts and reusable procedural skills used by
FeatureHospital.  A runtime request at stage \(s\) is assembled as
\begin{equation}
    P_s =
    \bigl[
        I_s;\,
        K_s;\,
        C_s(\mathcal{D});\,
        O_s
    \bigr],
\end{equation}
where \(I_s\) is the stage instruction, \(K_s\) is static skill knowledge,
\(C_s(\mathcal{D})\) is dataset-specific runtime context, and \(O_s\) is the
output contract.  The boxes below report the static instructions and skills.
Large dataset-specific JSON payloads are represented by named input slots,
because their values change with the dataset but their schemas do not.

We use \emph{prompt} for the role, task, constraints and output
schema sent to the LLM.  We use \emph{skill} for a reusable procedural
capability with an applicability condition, an execution procedure, tool
interactions, a stopping condition, and validation rules.  All six specialist
skills use dataset-independent loss catalogs; the LLM selects only among
implemented entries and cannot synthesize a new loss.

\subsection{Dataset Analysis}
\label{app:dataset_analysis}

Dataset analysis first computes deterministic statistics from
\((\mathcal{X}_{tr},Y_{tr})\).  Rule-confirmed findings are then passed to the LLM only
for concise diagnostic wording.  Thus, the LLM does not decide whether an
abnormality exists at this stage.

\label{app:dataset_analysis_prompt}

\begin{casebox}{Dataset Issue-Card Prompt}
\small
\textbf{User-message opening.}
``You are generating category-level dataset issue cards for a multi-label
feature selection triage system.''  The message states that the supplied
findings are already rule-confirmed and grouped by category, with fixed
\FHCode{finding_role}, metrics, triggered rules, affected items, and
explanations.

\textbf{Exact runtime payload fields.}
After the static instruction, the prompt appends an \FHCode{Input} JSON object
containing \FHCode{dataset_name}, outer-training \FHCode{basic_info}, and the
full deterministic \FHCode{issue_cards}.  Each card includes its identifiers,
category, abnormality summary, confirmed issue tags, and complete protected
finding evidence.

\textbf{Task.}
\begin{enumerate}
    \item Write one short \FHCode{diagnosis_sentence} for every confirmed
    finding.
    \item Write one \FHCode{category_diagnostic_summary} for each card.
    \item Write \FHCode{handoff_notes} indicating which canonical problem
    department may need the card.
\end{enumerate}

\textbf{Constraints.}
The prompt explicitly says: do not decide whether an issue exists; remove an
issue; add an issue tag; change \FHCode{finding_role}, level, severity,
confidence, affected items, triggered rules, evidence, or metric dictionaries;
or recommend modules, losses, or weights.  It also prohibits selected
features, rankings, training, baselines, classifiers, and test metrics.  Only
concise diagnostic text may change.

\textbf{Output.}
Return JSON only.  The top-level field is \FHCode{issue_cards}; each item
contains \FHCode{card_id}, \FHCode{category_diagnostic_summary},
\FHCode{handoff_notes}, and a list of
\(\{\FHCode{issue_tag},\FHCode{diagnosis_sentence}\}\) pairs.

\textbf{Repair and failure behavior.}
A schema or protected-field violation causes a bounded retry whose user
message includes the previous validation error.  Under the public protocol's
default \FHCode{fail} policy, exhausted retries terminate that seed.  An
explicit \FHCode{fallback} mode exists only for diagnostic runs and is not used
for reported LLM results.
\end{casebox}

\label{app:dataset_analysis_skills}

\begin{casebox}{Deterministic Statistical Profiling Skill}
\small
\textbf{Applicability.}
Run for every input dataset before any LLM decision.  All statistics are
computed from the currently available analysis partition.

\textbf{Procedure and tools.}
\begin{enumerate}
    \item \textbf{Basic-scale profiler:} compute sample, feature, label, and
    view counts; feature-to-sample ratio; label-to-sample ratio; and top-\(k\)
    budget indicators.
    \item \textbf{Label-distribution profiler:} compute per-label frequency,
    positive counts, imbalance ratio, Gini coefficient, label cardinality,
    label density, and rare-label statistics.
    \item \textbf{Feature-quality profiler:} compute missing and non-finite
    rates, zero rates, variance, unique-value ratios, scale heterogeneity, and
    outlier indicators.
    \item \textbf{Feature-redundancy profiler:} compute absolute feature
    correlations, high-correlation edge density, maximum local correlation,
    duplicate rate, effective rank, and redundant-cluster statistics.
    \item \textbf{Label-dependency profiler:} compute label correlations,
    Jaccard overlap, positive and negative dependency edges, components,
    clusters, and near-duplicate label groups.
    \item \textbf{Multi-view profiler:} compute view sizes, sparsity,
    missingness, scale heterogeneity, relevance, and reliability differences.
    \item \textbf{Local-structure profiler:} construct a fixed neighborhood
    graph and compute local label disagreement, consistency, hubness, and
    feature-label neighborhood alignment.
    \item \textbf{Feature-label profilers:} compute relevance strength,
    specificity, concentration, weak-signal indicators, subspace alignment,
    and optional MI/CMI complementarity statistics.
\end{enumerate}

The resulting measurements are evaluated by the fixed diagnostic rule
catalog.  A triggered rule produces a finding with a fixed issue tag,
finding role, severity, confidence, evidence metrics, and affected items.
Nine category cards organize these findings: basic scale, label distribution,
feature quality, feature redundancy, label dependency, multi-view structure,
local structure, feature-label relevance, and feature-label space structure.

\textbf{Termination.}
Stop after every applicable profiler has either produced its statistics or
explicitly recorded that the required data structure is unavailable.

\textbf{Validation.}
Array dimensions must agree with \((n,d,L)\); view slices must cover valid
feature indices; statistics used by a triggered rule must be present and
finite when defined; and no LLM-generated issue may enter the confirmed
finding set.
\end{casebox}

\begin{casebox}{Issue-Card Rendering Skill}
\small
\textbf{Applicability.}
Apply after deterministic profiling has produced grouped, rule-confirmed
findings.

\textbf{Procedure.}
Preserve every protected field, ask the LLM only for diagnostic sentences,
category summaries, and handoff notes, parse the returned JSON, and merge
only those textual fields into the deterministic cards.

\textbf{Tool interaction.}
The skill consumes the statistical profile and confirmed-finding catalog.  It
does not call an optimizer, classifier, feature selector, or evaluation tool.

\textbf{Termination and validation.}
Accept the response only if all returned card and issue identifiers already
exist and all protected evidence fields are unchanged.  Invalid JSON or a
semantic mismatch triggers a bounded repair attempt.  The deterministic card
may replace the LLM result only when explicit diagnostic fallback mode is
enabled; under the reported FHCode{fail} policy, exhausted retries terminate
the seed.
\end{casebox}

\subsection{Triage}
\label{app:triage}

\label{app:triage_prompt}

\begin{casebox}{Triage-Doctor Prompt}
\small
\textbf{User-message opening.}
``You are a department-routing triage agent for a dataset-intrinsic diagnosis
system.''  The prompt states that every issue-card finding is already marked
\FHCode{abnormal}, \FHCode{context}, or \FHCode{beneficial} and carries its
metrics, affected items, and diagnostic summary.

\textbf{Task.}
\begin{enumerate}
    \item Map confirmed problems to the allowed specialist departments.
    \item Assign one of \FHCode{activate}, \FHCode{absorbed},
    \FHCode{supporting}, \FHCode{context_only}, or \FHCode{inactive}.
    \item Activate a department only when it represents an independent
    downstream problem.
    \item If another activated department sufficiently explains a problem,
    record \FHCode{absorbed_by}; retain background findings in
    \FHCode{context_only}.
    \item Copy only relevant core metrics and explain every routing decision.
\end{enumerate}

\textbf{Allowed departments.}
\FHCode{LABEL_IMBALANCE}, \FHCode{FEATURE_REDUNDANCY},
\FHCode{VIEW_QUALITY_IMBALANCE}, \FHCode{LABEL_DEPENDENCY},
\FHCode{FEATURE_QUALITY_DEFECT}, and
\FHCode{LOCAL_LABEL_INCONSISTENCY}.

\textbf{Exact runtime payload fields.}
The appended JSON contains \FHCode{dataset_name}, outer-training
\FHCode{basic_info}, all \FHCode{issue_cards},
\FHCode{department_responsibilities}, and
\FHCode{routing_status_values}.  No precomputed routing answer is included.
The model must infer each disposition from the cards, the fixed six-department
responsibility catalog, and the routing rules.

\textbf{Constraints.}
Do not invent a department, issue tag, metric, module, loss, or weight.  Do
not prescribe an objective or use primary/secondary priority.  Source issues
and metrics must be copied from the issue cards.  The prompt additionally
states that \FHCode{FEATURE_BUDGET_PRESSURE} is emitted deterministically as a
global constraint and must not be assigned to a department.  Return JSON only.

\textbf{Absorption instructions.}
Minor-label signal weakness and label-signal imbalance are routed to
\FHCode{LABEL_IMBALANCE}; conditional relevance dependence is routed to
\FHCode{LABEL_DEPENDENCY}; generic weak signal, predictability weakness, or
noncompactness is routed only through a confirmed root cause owned by an
implemented department, otherwise it remains context.  The prompt also treats
\FHCode{MULTI_VIEW_STRUCTURE} as context unless a view abnormality exists,
\FHCode{LABEL_SUBSPACE_NONCOMPACTNESS} as context unless directly usable by
an implemented department, and \FHCode{LOCAL_LABEL_CONSISTENCY} as
context/beneficial rather than local inconsistency.

\textbf{Output.}
The LLM is explicitly asked for \FHCode{department_cases},
\FHCode{activated_departments}, \FHCode{absorbed_departments}, and
\FHCode{context_only}.  Each department case records
\FHCode{department}, \FHCode{routing_status}, \FHCode{source_cards},
\FHCode{source_issues}, \FHCode{absorbed_by}, \FHCode{absorbed_issues},
\FHCode{supporting_context_issues}, \FHCode{beneficial_context_issues},
\FHCode{problem_strength}, \FHCode{confidence}, \FHCode{core_metrics},
\FHCode{affected_items_summary}, and \FHCode{routing_reason}.

\textbf{Validation, repair, and deterministic augmentation.}
The response is checked against the issue cards and registry.  A rejected
response is retried with the prior validation error and a JSON-only repair
instruction.  After acceptance, the runtime deterministically attaches
\FHCode{global_constraints} and \FHCode{routing_coverage}.  Under the public
protocol's default \FHCode{fail} policy, exhausted retries terminate that seed.
The independent deterministic router is used only when a diagnostic run
explicitly requests deterministic mode or fallback behavior.
\end{casebox}

\label{app:triage_skills}

\begin{casebox}{Evidence-to-Department Routing Skill}
\small
\textbf{Applicability.}
Apply once the issue-card set is complete.

\textbf{Canonical department registry.}

\noindent\FHCode{LABEL_IMBALANCE}\\
\hspace*{1em}-- \textbf{Treatment:} Uneven label frequency or signal and
insufficient rare-label protection.
\par\smallskip

\noindent\FHCode{FEATURE_REDUNDANCY}\\
\hspace*{1em}-- \textbf{Treatment:} Duplicated feature information and
dominant local redundant groups.
\par\smallskip

\noindent\FHCode{VIEW_QUALITY_IMBALANCE}\\
\hspace*{1em}-- \textbf{Treatment:} Cross-view differences in quality,
sparsity, scale, size, or allocation.
\par\smallskip

\noindent\FHCode{LABEL_DEPENDENCY}\\
\hspace*{1em}-- \textbf{Treatment:} Positive, negative, clustered, or
near-duplicate label relations.
\par\smallskip

\noindent\FHCode{FEATURE_QUALITY_DEFECT}\\
\hspace*{1em}-- \textbf{Treatment:} Missing, non-finite, near-constant,
outlier-driven, or unstable features.
\par\smallskip

\noindent\FHCode{LOCAL_LABEL_INCONSISTENCY}\\
\hspace*{1em}-- \textbf{Treatment:} Misalignment between feature-space
neighborhoods and label-set similarity.

\textbf{Procedure.}
\begin{enumerate}
    \item Separate abnormal findings from contextual and beneficial findings.
    \item Send the original issue cards, outer-training basic information,
    Department responsibilities, and allowed statuses to the LLM without a
    precomputed routing answer.
    \item Let the LLM construct a disposition within the six-department
    registry using the stated independence and absorption logic.
    \item Validate every returned source issue, context item, metric, status,
    and absorption reference against the issue cards and registry.
    \item Treat top-\(k\) feature-budget pressure as a deterministic global
    constraint rather than a specialist department, and attach routing
    coverage after the LLM response is accepted.
    \item If all LLM repairs fail under the reported protocol, fail that seed
    rather than changing the decision source silently.
\end{enumerate}

\textbf{Absorption knowledge.}
Weak feature-label signal or budget pressure concentrated on rare labels may
support \FHCode{LABEL_IMBALANCE}; \FHCode{MULTI_VIEW_STRUCTURE} alone is
context; \FHCode{LOCAL_LABEL_CONSISTENCY} is beneficial context rather than
local inconsistency; and generic alignment or subspace findings remain
context unless an implemented department can directly use them.

\textbf{Termination and validation.}
Stop when every finding is activated, absorbed, supporting, a global
constraint, or explicit context.  An activated case must contain at least
one abnormal source issue.  Every department identifier must belong to the
canonical registry, every \FHCode{absorbed_by} reference must point to an
activated department, and all referenced metrics must exist in the source
cards.  The derived \FHCode{activated_departments} and
\FHCode{absorbed_departments} lists must agree with the corresponding case
statuses.
\end{casebox}

\subsection{Consultation}
\label{app:consultation}

In the strict runner, each Doctor prompt is regenerated from the outer-training
partition.  Its dataset summary uses the 20\% consultation budget with
\FHCode{max(1,floor(0.20*d))} features.  Before constructing Doctor inputs only,
\FHCode{yeast.mat} and \FHCode{emotions.mat} are transformed featurewise by
three-bin equal-width discretization; the other datasets retain their loaded
feature values.  This as-run preprocessing therefore affects the Specialist
Consultation payload but is not applied to the earlier Dataset Analysis
payload or the later Pharmacist mechanics probe.

For notation, let \(X\in\mathbb{R}^{n\times d}\) be the feature matrix,
\(Y\in\{0,1\}^{n\times L}\) the label matrix, \(R\in\mathbb{R}^{d\times L}\)
the feature-label relevance matrix, and \(z\in[0,1]^d\) the continuous
selection mask.  Most structural losses use the budget-comparable mask
\begin{equation}
    z_j^{\mathrm{eff}}
    =
    \frac{z_j}{\sum_{r=1}^{d} z_r+\epsilon}\,k,
\end{equation}
where \(k\) is the requested number of selected features.

\label{app:consultation_prompt}

\begin{casebox}{Shared Department-Doctor Prompt}
\small
\textbf{User-message opening.}
The shared template says: ``You are an expert in
\(\langle\)department domain\(\rangle\).  You are acting as the doctor for the
\(\langle\)department name\(\rangle\) department.''  It then appends that
department's task description.

\textbf{Exact runtime sections.}
The user message contains, in order:
\begin{itemize}
    \item full \FHCode{Domain knowledge};
    \item outer-training \FHCode{Dataset summary};
    \item the \FHCode{Activated department case}, including its source cards,
    issues, strength, confidence, metrics, and routing reason;
    \item the \FHCode{Implemented capability catalog}, containing the
    available losses and tools;
    \item \FHCode{Allowed parameter ranges};
    \item the \FHCode{Required JSON schema}; and
    \item \FHCode{Previous validation error, if any}.
\end{itemize}

\textbf{General rules exactly conveyed.}
Return JSON only; do not invent losses, metrics, modules, formulas,
departments, or parameters; and use only implemented catalog capabilities.
The strict run set \FHCode{max_candidates=3}, so the prompt explicitly asked
for one to three compact candidate objectives.  It instructs the Doctor to
order \FHCode{candidate_objectives} from most to least recommended and states
that \FHCode{candidate_objectives[0]} must be the final recommended Department
prescription because the pipeline selects it directly.  Alternatives are
included only for genuine evidence-supported mechanism trade-offs.

\textbf{Output.}
Return \FHCode{candidate_objectives}; each candidate contains a unique ID,
rationale, \FHCode{enabled_loss_names}, and parameter values within the
catalog bounds.

\textbf{Actual candidate-selection behavior.}
After schema validation, the runner selects the explicitly ranked first
candidate and records \FHCode{policy=first_ranked_candidate}, the decision
source, and candidate count.  It then converts that candidate into the
Department prescription.  No hidden candidate scoring, predictive validation,
or automatic search is performed.

\textbf{Failure behavior.}
All six Doctor selectors use bounded schema-repair attempts.  On exhausted
retries, the selector raises an error; the public protocol's default
\FHCode{fail} policy terminates that seed.  A deterministic replacement is
available only through an explicitly selected diagnostic mode or fallback
policy.
\end{casebox}

\label{app:consultation_skills}

\begin{casebox}{Shared Specialist-Consultation Skill}
\small
\textbf{Applicability.}
Instantiate this skill independently for every activated department.

\textbf{Procedure.}
\begin{enumerate}
    \item Read only the routing evidence assigned to the department.
    \item Perform a department-specific differential diagnosis and reject
    explanations that belong to peer departments.
    \item Match confirmed symptoms to implemented loss mechanisms, considering
    benefit, side effect, degeneracy, and interaction risk.
    \item Produce one to three schema-valid candidate prescriptions and valid
    parameters in descending recommendation order; the runtime selects the
    declared first choice without empirically evaluating alternatives.
    \item Attach the exact formula, required tools, required inputs, intended
    effect, risks, and tuning bounds to every selected loss.
\end{enumerate}

\textbf{Termination and validation.}
Stop after one to three schema-valid candidates are produced.  Every enabled
loss and parameter must occur in the department catalog; all values must lie
inside the declared ranges; and no peer-department loss may be introduced.
On a validation failure, the same template is resent with the error populated
in \FHCode{Previous validation error, if any}.
\end{casebox}

\begin{casebox}{LABEL\_IMBALANCE Skill}
\small
\textbf{Goal.}
Prevent frequent labels from dominating supervised feature relevance and
protect rare labels under a limited top-\(k\) budget.

\textbf{Applicability and differential diagnosis.}
Use label-frequency skew, rare-label counts, weak minor-label relevance, and
label-wise contribution imbalance.  Do not treat view allocation, feature
duplication, or global label-graph structure as this department's primary
problem.

\textbf{Loss catalog.}
\begin{enumerate}
    \item \FHCode{label_weighted_relevance_loss}:
    \[
        \mathcal{L}_{\mathrm{lwr}}
        =-\frac{1}{dL}\sum_{j=1}^{d}z_j
        \sum_{\ell=1}^{L}w_\ell R_{j\ell}.
    \]
    It is the first-line response to head-label domination.  Rare-label
    weights are clipped by \FHCode{w_max} to avoid amplifying estimates based
    on very few positives.

    \item \FHCode{rare_label_coverage_loss}:
    \[
        \mathcal{L}_{\mathrm{rare}}
        =\frac{1}{|\mathcal{R}|}\sum_{\ell\in\mathcal{R}}
        \left[
        \rho\!\!\sum_{j\in\operatorname{Top}_m(R_{:\ell})}\!\!R_{j\ell}
        -\sum_j z_j^{\mathrm{eff}}R_{j\ell}
        \right]_{+}.
    \]
    Use it when rare labels and budget pressure are both supported.  A
    persistently zero loss is treated as nonbinding rather than evidence of a
    solved problem.

    \item \FHCode{label_contribution_balance_loss}:
    \[
        \mathcal{L}_{\mathrm{bal}}
        =\operatorname{Var}_{\ell}
        \left(
        \frac{w_\ell\sum_jz_jR_{j\ell}}
        {\operatorname{mean}_{q}
        (w_q\sum_jz_jR_{jq})+\epsilon}
        \right).
    \]
    Use a conservative dose when selected-feature contribution is strongly
    concentrated; excessive equalization may suppress genuinely strong
    labels.

    \item \FHCode{budget_penalty}:
    \[
        \mathcal{L}_{\mathrm{budget}}
        =\frac{(\sum_jz_j-k)^2}{k^2}.
    \]
    This controls continuous mask mass when top-\(k\) pressure is present.
\end{enumerate}

\textbf{Tools.}
The losses consume label frequencies, \(R\), \(z\), and \(k\); no additional
department-specific tool is required.

\textbf{Validation.}
Check that rare-label protection does not collapse overall relevance, that a
coverage term is active when credited, and that the prescription does not
attempt to solve view concentration or redundancy directly.
\end{casebox}

\begin{casebox}{FEATURE\_REDUNDANCY Skill}
\small
\textbf{Goal.}
Prevent correlated or near-duplicate features from wasting the top-\(k\)
budget while retaining correlated features that support different labels.

\textbf{Applicability.}
Use high local maximum correlations, selected high-correlation pair
fractions, or large redundant clusters.  Broad edge density favors pairwise
control; sparse but large local clusters favor cluster quota control.

\textbf{Tool.}
\FHCode{sparse_feature_redundancy_graph_builder} constructs a sparse
feature-feature graph, label-profile-aware edge weights, redundant clusters,
internal-correlation statistics, and selected-subset diagnostics.  Its output
is cached and shared by both losses.

\textbf{Loss catalog.}
\begin{enumerate}
    \item \FHCode{label_aware_pairwise_redundancy_loss}:
    \[
    \begin{aligned}
        A_{ij}
        &=|\operatorname{corr}(x_i,x_j)|^{p_f}
        \bigl[\cos(w\odot R_{i:},w\odot R_{j:})\bigr]_{+}^{p_\ell},
        \\
        \mathcal{L}_{\mathrm{pair}}
        &=\frac{\sum_{(i,j)\in E}A_{ij}
        z_i^{\mathrm{eff}}z_j^{\mathrm{eff}}}
        {\sum_{(i,j)\in E}A_{ij}+\epsilon}.
    \end{aligned}
    \]
    It is applicable when highly correlated selected pairs also have similar
    label-relevance profiles.

    \item \FHCode{redundant_cluster_quota_loss}:
    \[
        m_c=\sum_{j\in C_c}z_j^{\mathrm{eff}},\quad
        q_c=\operatorname{clip}\!\left(
        \left\lceil\rho_c^{q}k|C_c|/d\right\rceil,
        q_{\min},q_{\max}\right),
    \]
    \[
        \mathcal{L}_{\mathrm{quota}}
        =\frac{\sum_c
        \log(1+|C_c|)\bar r_c[m_c-q_c]_{+}^{2}}
        {\sum_c\log(1+|C_c|)\bar r_c+\epsilon}.
    \]
    It is preferred when a few large local redundant clusters can dominate
    selection.
\end{enumerate}

\textbf{Validation.}
The graph must remain sparse, penalties must be label-profile aware when
available, and quotas must not remove correlated features merely because
they serve rare labels or distinct views.
\end{casebox}

\begin{casebox}{VIEW\_QUALITY\_IMBALANCE Skill}
\small
\textbf{Goal.}
Make relevance comparable across views and prevent harmful allocation caused
by view size, sparsity, scale, or reliability differences without enforcing
uniform quotas.

\textbf{Applicability.}
Use confirmed view-level quality heterogeneity, view-specific sparsity,
feature-count imbalance, or selected-feature concentration.  Feature-level
defects remain the responsibility of \FHCode{FEATURE_QUALITY_DEFECT}.

\textbf{Tool.}
\FHCode{view_quality_profiler} computes view IDs and slices, sizes, zero and
missing rates, scale heterogeneity, relevance, reliability, quality,
quality-aware target allocation, and selected-view diagnostics.  One profile
is reused by all view losses.

\textbf{Loss catalog.}
\begin{enumerate}
    \item \FHCode{view_normalized_relevance_loss}.  Define
    \(s_j=\sum_\ell u_\ell R_{j\ell}\),
    \(a_v=\operatorname{meanTop}_m\{s_j:j\in v\}+\epsilon\),
    \(\tilde s_j=\operatorname{clip}(s_j/a_{v(j)},0,c)\), and
    \(q_v=(\eta Q_v+(1-\eta)U_v)^p\).  Then
    \[
        \mathcal{L}_{\mathrm{vnr}}
        =-\frac{\sum_jz_j^{\mathrm{eff}}q_{v(j)}\tilde s_j}
        {k\,\operatorname{mean}_j(q_{v(j)}\tilde s_j)+\epsilon}.
    \]
    Use it when raw relevance scores are not comparable across views.

    \item \FHCode{adaptive_view_allocation_loss}.  Let the normalized target
    \(\pi_v\) mix quality, relevance, square-root view size, and a uniform
    prior; \(t_v=k\pi_v\), with tolerance interval
    \([l_v,u_v]\).  For \(m_v=\sum_{j\in v}z_j^{\mathrm{eff}}\),
    \[
        \mathcal{L}_{\mathrm{alloc}}
        =\frac{\sum_v
        \alpha_u[m_v-u_v]_{+}^{2}
        +\alpha_l[l_v-m_v]_{+}^{2}}
        {k^2+\epsilon}.
    \]
    Use only for harmful concentration and retain a non-uniform,
    quality-aware target.

    \item \FHCode{view_coverage_floor_loss}:
    \[
        \mathcal{L}_{\mathrm{floor}}
        =\frac{\sum_{v:e_v=1}[f_v-m_v]_{+}^{2}}
        {\sum_{v:e_v=1}f_v^2+\epsilon},
    \]
    where eligibility \(e_v\) requires sufficient view quality, relevance,
    and budget, and \(f_v\) is a bounded fraction of the target.  Use only
    when eligible views are excluded and \(k\) can support multi-view
    coverage.
\end{enumerate}

\textbf{Validation.}
Do not credit an inactive floor, do not force equal view counts, and check
that normalization does not amplify a low-quality sparse view.
\end{casebox}

\begin{casebox}{LABEL\_DEPENDENCY Skill}
\small
\textbf{Goal.}
Preserve useful positive, negative, clustered, and near-duplicate label
relations without replacing feature selection with a heavy label predictor.

\textbf{Applicability.}
Use confirmed dependency edges, label communities, mutually exclusive
labels, or near-duplicate labels.  Global label relations are distinct from
sample-neighborhood inconsistency.

\textbf{Tools.}
\FHCode{label_dependency_graph_builder} constructs positive and negative
edges, edge weights, near-duplicate groups, and graph components.
\FHCode{near_duplicate_label_deweighting} softly reduces repeated label
counting and is not a standalone loss.

\textbf{Loss catalog.}
\begin{enumerate}
    \item \FHCode{label_graph_smoothed_relevance_loss}.  With
    \(P^+=\operatorname{rowNorm}(A^+)\),
    \[
    \begin{aligned}
        R^{g}_{j\ell}
        &=(1-\alpha)R_{j\ell}
        +\alpha\sum_mP^+_{\ell m}R_{jm},\qquad
        \\
        \mathcal{L}_{\mathrm{graph}}
        &=-\frac{\sum_jz_j^{\mathrm{eff}}\sum_\ell u_\ell R^g_{j\ell}}
        {k\sum_\ell u_\ell+\epsilon}.
    \end{aligned}
    \]
    Conservative smoothing lets weak labels borrow signal while limiting
    head-label propagation.

    \item \FHCode{positive_dependency_cocoverage_loss}:
    \[
        \mathcal{L}_{\mathrm{co}}
        =-\frac{\sum_{(\ell,m)\in E^+}A^+_{\ell m}
        \sum_jz_j^{\mathrm{eff}}R_{j\ell}R_{jm}}
        {\sum_{(\ell,m)\in E^+}A^+_{\ell m}+\epsilon}.
    \]
    Use when reliable positive edges should be jointly covered.

    \item \FHCode{label_cluster_coverage_loss}.  For
    \(R^c_{jc}=|C_c|^{-1}\sum_{\ell\in C_c}R^g_{j\ell}\),
    \(a_c=\sum_jz_j^{\mathrm{eff}}R^c_{jc}\), and a top-\(m\)
    target \(\tau_c\),
    \[
        \mathcal{L}_{\mathrm{cluster}}
        =\frac{\sum_c\rho_c[\tau_c-a_c]_{+}}
        {\sum_c\rho_c+\epsilon}.
    \]
    It protects smaller label communities but competes for top-\(k\) capacity.

    \item \FHCode{negative_dependency_separation_loss}:
    \[
        \mathcal{L}_{\mathrm{neg}}
        =\frac{\sum_{(\ell,m)\in E^-}A^-_{\ell m}
        \sum_jz_j^{\mathrm{eff}}
        [R^{\pm}_{j\ell}R^{\pm}_{jm}-\delta]_{+}}
        {\sum_{(\ell,m)\in E^-}A^-_{\ell m}+\epsilon}.
    \]
    This is a low-dose guard against same-direction support for mutually
    exclusive labels.
\end{enumerate}

\textbf{Validation.}
Positive smoothing must not propagate head-label dominance, negative edges
must have adequate support, and overlapping relevance rewards must be exposed
to the pharmacist for de-duplication.
\end{casebox}

\begin{casebox}{FEATURE\_QUALITY\_DEFECT Skill}
\small
\textbf{Goal.}
Keep missing-heavy, non-finite, near-constant, outlier-driven, or statistically
unstable features from occupying the selected subset.

\textbf{Applicability and boundary.}
Use feature-level defects and relevance unreliability.  Redundancy concerns
relationships between features, while view-quality imbalance concerns
aggregate view behavior.  High zero rate alone is not a hard-invalid
condition.

\textbf{Tools.}
The catalog presents \FHCode{invalid_feature_filter} as a conservative
pre-optimization capability for identifying all-missing, unusably non-finite,
constant, or single-unique-value features and exposing a valid-feature mask.
The
\FHCode{feature_quality_profiler} computes per-feature missingness,
non-finite rate, zero rate, variance, unique ratio, outlier rate, bootstrap
relevance stability, robust-relevance consistency, defect score, and quality
score.  The resulting arrays are reusable across candidate losses.

\textbf{Loss catalog.}
\begin{enumerate}
    \item \FHCode{defective_feature_suppression_loss}:
    \[
        \mathcal{L}_{\mathrm{defect}}
        =\frac{1}{k+\epsilon}\sum_j
        z_j^{\mathrm{eff}}
        \operatorname{clip}(d_j,0,1)^p.
    \]
    This is the safest first-line soft prior when a high feature defect score
    is confirmed.

    \item \FHCode{unstable_relevance_penalty_loss}.  With bootstrap aggregate
    score \(s_{bj}=\sum_\ell u_\ell R^{(b)}_{j\ell}\),
    \[
    \begin{aligned}
        h_j&=\operatorname{clip}\!\left(
        \frac{\operatorname{std}_b(s_{bj})}
        {|\operatorname{mean}_b(s_{bj})|+\epsilon},0,c\right),
        \\
        \mathcal{L}_{\mathrm{unstable}}
        &=\frac{1}{k+\epsilon}\sum_jz_j^{\mathrm{eff}}(h_j/c)^p.
    \end{aligned}
    \]
    Use a small dose when relevance instability is confirmed; bootstrap
    labels with insufficient positives are excluded.

    \item \FHCode{robust_relevance_consistency_loss}.  Let
    \(a_j=\sum_\ell u_\ell|R^{\mathrm{raw}}_{j\ell}|\) and
    \(b_j=\sum_\ell u_\ell|R^{\mathrm{rob}}_{j\ell}|\).  Then
    \[
        q_j=\left[
        \frac{a_j-b_j}{a_j+\epsilon}-\tau
        \right]_{+},\qquad
        \mathcal{L}_{\mathrm{robust}}
        =\frac{1}{k+\epsilon}\sum_jz_j^{\mathrm{eff}}q_j^p.
    \]
    Use only when outlier-driven or robust-inconsistent relevance is
    supported.
\end{enumerate}

\textbf{Validation.}
The Doctor knowledge states that hard-invalid selected count should be zero
and that soft penalties must not treat sparsity alone as a defect or erase
rare-label event features merely because their relevance estimates are
noisier.  In the historical strict executor, however, the hard-invalid mask
was recorded and assigned maximal defect risk but was not imposed as a hard
constraint on the final \FHCode{argsort}.  Zero hard-invalid selections was
therefore a requested diagnostic target, not a mechanical guarantee of the
as-run optimizer.
\end{casebox}

\begin{casebox}{LOCAL\_LABEL\_INCONSISTENCY Skill}
\small
\textbf{Goal.}
Prefer features that make feature-space neighborhoods label-consistent or
locally discriminative when nearby samples have substantially different
label sets.

\textbf{Applicability and boundary.}
The triage evidence, rather than a single hard-coded issue tag, determines
activation.  The mismatch must be an independent sample-neighborhood problem,
not merely a graph that can be constructed, global label dependency, a
feature-quality artifact, or a view-scale artifact.

\textbf{Tool.}
\FHCode{local_label_consistency_profiler} robustly scales training features,
builds one fixed training-only \(k\)-NN graph, separates reliable
label-consistent and label-inconsistent edges, and computes edge reliability,
local disagreement, rare-label dilution, neighborhood alignment, and
per-feature local discriminative scores.  The graph is not rebuilt during
mask optimization.

\textbf{Loss catalog.}
\begin{enumerate}
    \item \FHCode{local_discriminative_relevance_loss}.  With inconsistent
    edges \(E_{\mathrm{inc}}\), consistent edges \(E_{\mathrm{con}}\), and
    reliability \(a_{uv}\),
    \[
    \begin{aligned}
        s_j^{\mathrm{loc}}
        =
        &\operatorname{mean}_{(i,n)\in E_{\mathrm{inc}}}
        a_{in}|x_{ij}-x_{nj}|
        \\
        &-\beta
        \operatorname{mean}_{(i,p)\in E_{\mathrm{con}}}
        a_{ip}|x_{ij}-x_{pj}|,
    \end{aligned}
    \]
    \[
        \mathcal{L}_{\mathrm{local}}
        =-\frac{1}{k+\epsilon}\sum_j
        z_j^{\mathrm{eff}}\widetilde{s}_j^{\mathrm{loc}}.
    \]
    This precomputed feature-level score is the first-line local treatment.

    \item \FHCode{local_inconsistent_neighbor_separation_}
    
    \FHCode{loss}.  For the
    selected-space distance
    \[
        d_z(i,n)=\frac{1}{k+\epsilon}
        \sum_jz_j^{\mathrm{eff}}
        \operatorname{NormDist}(x_{ij},x_{nj}),
    \]
    \[
        \mathcal{L}_{\mathrm{sep}}
        =\operatorname{mean}_{(i,n)\in E_{\mathrm{inc}}}
        a_{in}[\gamma-d_z(i,n)]_{+}^{2}.
    \]
    This direct separation term is optional and low-dose because an
    inconsistent edge may reflect ambiguity or label noise.
\end{enumerate}

\textbf{Validation.}
All graph construction uses training samples only; edge count is bounded;
unreliable edges are filtered or downweighted; and local compactness must not
collapse samples that share only a frequent label.
\end{casebox}

\subsection{Objective Construction}
\label{app:obj_const}

\label{app:obj_const_prompt}

\begin{casebox}{Pharmacist Objective-Construction Prompt}
\small
\textbf{Base user-message role.}
The message identifies the model as the Pharmacist for a multi-view
multi-label feature-selection system.  Department Doctors have examined
activated Department cases and produced prescriptions; the Pharmacist must merge
them into one compact, conflict-aware global objective without re-diagnosing
the dataset or empirically validating prediction.

\textbf{Base role boundary.}
The Pharmacist may use only confirmed dataset context, Doctor-selected losses,
implemented formulas, tools, inputs, bounds, and mandatory global guardrails.
It must not invent an issue, loss, formula, module, tool, metric, dataset, or
department; run training, testing, KNN/classifier evaluation, or automatic
tuning; reconsider a Doctor-disabled loss; or emit a loss absent from
\FHCode{department_prescriptions[*].selected_}

\FHCode{losses}.  Every Doctor-selected
loss must occur exactly once as enabled or disabled.  Global guardrails must
be copied unchanged and must not be duplicated among department losses.

\textbf{Base fusion rules.}
The prompt asks for one main relevance backbone, complementary treatments for
distinct confirmed problems, conservative handling of top-\(k\) competition,
de-duplication of similar relevance rewards, preservation of tool dependencies
and Doctor parameter bounds, fresh global roles, and a shared tool-execution
plan.  It permits \FHCode{keep}, \FHCode{downweight}, \FHCode{upweight},
\FHCode{merge}, and \FHCode{disable}.

\textbf{Exact runtime sections and order.}
The rendered user prompt contains:
\begin{enumerate}
    \item \FHCode{[general description]}, \FHCode{[role boundary]},
    \FHCode{[domain knowledge]}, \FHCode{[operating rules]}, and
    \FHCode{[reasoning checklist]};
    \item full outer-training \FHCode{[dataset_pharmacist_context]};
    \item \FHCode{[required_global_guardrails]};
    \item all \FHCode{[department_prescriptions]};
    \item the inserted \FHCode{[final-dose contract]};
    \item a \FHCode{[doctor-derived provisional probe objective]} containing
    every Doctor-selected loss at the first candidate's initial settings;
    \item \FHCode{[low-cost pharmacodynamic evidence]}, including term
    activity, loss values, gradient norms, effective pressure, cosine and
    top-\(k\) overlap, mechanism sensitivity, counterfactual micro-probes,
    cache information, and \FHCode{compact_decision_clues};
    \item the fourteen-step \FHCode{[evidence-guided decision procedure]}; and
    \item \FHCode{[required JSON schema]}.
\end{enumerate}

\textbf{Final-dose contract.}
The base prompt and inserted evidence instructions agree that there is no
later automatic tuning.  The retained schema names \FHCode{initial_weight}
and \FHCode{initial_parameters} denote the final operational dose consumed by
the deterministic optimizer.  Doctor-provided ranges are preserved only as
catalog provenance; the output schema contains no executable search space.

\textbf{Evidence-guided instructions.}
The inserted procedure tells the LLM to select exactly one primary relevance
backbone when available; use problem strength and Doctor rationale; avoid
mapping absolute gradient size directly to dose under Adam; inspect duplicate
and opposing pressure; read compact clues before detailed probes; obey
\FHCode{mechanical_preferred_status} and
\FHCode{conservative_weight_ceiling} subject to stated exceptions; inspect
mechanism sensitivity before a large dose change; retain distinct severe
structural treatment when justified; and account for every Doctor-selected
loss exactly once.  Probe endpoint changes are explicitly described as
objective-mechanics evidence, not predictive evidence.

\textbf{Output and runtime normalization.}
The required JSON contains \FHCode{global_loss_blueprint},
\FHCode{resolved_conflicts}, \FHCode{operational_weight_plan},
\FHCode{tool_execution_plan}, and
\FHCode{pharmacist_summary}.  The blueprint contains
\FHCode{global_guardrail_terms}, \FHCode{selected_terms}, and
\FHCode{disabled_terms}.  Before validation, the runner injects the required
global guardrails into the parsed response.  A failed response is retried with
the previous validation error and up to the first 6000 characters of the
invalid response; no deterministic Pharmacist fallback is used by this
strict runner.
\end{casebox}

\label{app:obj_const_skills}

\begin{casebox}{Low-Cost Pharmacodynamic Evidence Skill}
\small
\textbf{Applicability.}
Run once after department prescriptions are available and before the
pharmacist LLM call.  The strict runner creates one additional 70/30
diagnostic split inside the outer-training partition with diagnostic seed
zero, then uses only that inner-training part for the mechanics probe.  It
never uses the outer-test partition, a classifier, or a predictive metric.

\textbf{Reusable tool preparation.}
Construct the shared relevance and label-weight profiles, then construct and
cache each structural artifact only when a prescribed probe term requires it.
This includes the label graph, feature-redundancy graph and clusters, view
profile, feature-quality profile, and fixed local graph.  Compatible artifacts
are reused across all probe configurations.  Label-graph construction and
near-duplicate deweighting are skipped when no enabled label-dependency term
requests them.

\textbf{Procedure.}
\begin{enumerate}
    \item Build a provisional objective containing all doctor-selected terms
    at doctor-proposed doses.
    \item Run a short primary-backbone-only optimization probe.
    \item At the initial mask and probed mask, compute each term's unit loss,
    activity state, raw gradient norm, effective pressure relative to the
    primary term, gradient cosine with the primary, and promoted top-\(k\)
    overlap.
    \item Flag near-duplicate pressure when pairwise gradient cosine is high
    and opposing pressure when it is substantially negative.
    \item Probe core mechanism parameters at representative low and high
    values while reusing cached artifacts.
    \item Run short objective-only counterfactuals: halve or increase a term's
    dose, remove each non-primary term, vary core mechanism parameters, and
    add a doctor-selected term omitted by the provisional reference.
    \item Summarize target-loss change, primary-loss change, top-\(k\) overlap,
    and budget-mass error into \FHCode{compact_decision_clues}.
\end{enumerate}

\textbf{Interpretation boundary.}
These probes measure local objective mechanics, not predictive efficacy.
Absolute gradient magnitude is not a direct dosage rule under Adam; negative
cosine is a review flag rather than proof of harm; and an inactive floor can
become active later.

\textbf{Termination and validation.}
Stop after the bounded intervention list is exhausted.  Verify that every
probe uses training data only, cached artifacts have compatible signatures,
no predictive metric was computed, and each clue can be traced to a
doctor-selected term.
\end{casebox}

\begin{casebox}{Evidence-Guided Prescription-Fusion Skill}
\small
\textbf{Applicability.}
Apply after low-cost evidence has been attached to the confirmed dataset
context and doctor catalog.

\textbf{Decision procedure.}
\begin{enumerate}
    \item Select exactly one global primary relevance backbone unless no
    doctor prescribed a relevance term.
    \item Use problem strength and doctor rationale to determine therapeutic
    necessity.
    \item Prefer one representative when relevance terms exert near-duplicate
    pressure.
    \item Read \FHCode{compact_decision_clues} before detailed micro-probe
    records.  A positive endpoint change means that the short intervention
    reduced that objective term, not that it improved prediction.
    \item Follow \FHCode{mechanical_preferred_status} for a nonbinding term
    unless it is the sole treatment for a severe confirmed issue; any
    exception must be justified and conservatively dosed.
    \item Treat \FHCode{conservative_weight_ceiling} as a hard ceiling unless
    a safe local intervention changes the top-\(k\) set, improves the target
    endpoint by at least \(0.02\), and degrades the primary endpoint by no
    more than \(0.01\).
    \item Do not regard a dose increase that lowers only its own loss while
    leaving top-\(k\) unchanged as useful dosage evidence.
    \item Inspect mechanism sensitivity before making a large weight change.
    \item Retain distinct structural treatments for severe problems even when
    they oppose the primary relevance gradient, but remove unnecessary
    duplicate pressure and excessive top-\(k\) competition.
    \item If probes are neutral or mixed, use the confirmed diagnosis and a
    conservative doctor dose rather than retaining every provisional term.
\end{enumerate}

\textbf{Global role classes.}
Enabled terms are assigned one of \FHCode{primary},
\FHCode{regularizer}, \FHCode{supporting}, or \FHCode{guardrail}.  Loss
classes include relevance backbone, coverage/floor, redundancy/diversity,
feature quality, local structure, allocation, defensive separation, and
budget guardrail.  These roles organize the final objective; they do not
require a fixed number of auxiliary terms.

\textbf{Termination.}
Stop when the objective is compact, every doctor-selected loss has exactly
one status, all mandatory guardrails are copied, final weights and mechanism
parameters are set, and the reusable-tool order is complete.
\end{casebox}

\begin{casebox}{Objective Blueprint Validation Skill}
\small
\textbf{Schema validation.}
Every enabled term must copy its department, loss name, formula, required
inputs, tools, and parameter bounds from one doctor-selected entry.  Every
disabled term must likewise refer to one doctor-selected entry and state
whether it was disabled or merged.  No loss may occur twice.

\textbf{Semantic validation.}
The primary role is unique when a relevance backbone exists; disabled doctor
losses are not reconsidered; global guardrails are unchanged; final values
lie within catalog bounds and respect evidence-derived ceilings; and all
conflicts and exceptions have explicit reasons.

\textbf{Execution validation.}
The tool plan must topologically precede every dependent loss and build each
shareable artifact once.  The accepted blueprint is then passed directly to
deterministic mask optimization, where the final ranking is obtained from the
learned feature scores \(z\); no further LLM call or automatic tuning modifies
the prescription.
\end{casebox}

\section{Implementation Details}
\label{app:impl}

Given the constructed objective, we optimize a continuous feature mask
$\mathbf{z}\in[0,1]^d$ on each training partition. The mask is
parameterized as $\mathbf{z}=\operatorname{sigmoid}(\mathbf{a}/T)$,
where $\mathbf{a}$ denotes the trainable logits and $T=1.0$. The logits
are initialized using the relevance-based feature scores and optimized
with Adam for 300 iterations. The learning rate is set to $0.03$, with
$\beta_1=0.9$, $\beta_2=0.999$, and $\epsilon=10^{-8}$. No additional
hyperparameter tuning is performed after objective construction.

For each random split, the objective is optimized once using a maximum
feature budget of $20\%$ of the original features. Features are ranked
according to their learned mask values $z_j$. The feature subsets at
selection ratios
$\{2\%,4\%,6\%,8\%,10\%,12\%,14\%,16\%,18\%,20\%\}$
are obtained by taking prefixes of this ranking, where the number of
selected features is
$\max(1,\lfloor r d\rfloor)$ for selection ratio $r$.

We randomly divide each dataset into $70\%$ training data and $30\%$
test data using ten fixed random seeds from 0 to 9. Following the evaluation protocol of the compared methods, we
employ a multi-output $10$-nearest-neighbor classifier for prediction.
Training samples whose label vectors contain no positive label are
removed, and labels without positive training examples are excluded
from the corresponding split.

We report Average Precision (AP), macro-averaged AUC, Ranking Loss (RL),
and exact-match zero-one loss (ZL). For each random split, a metric is
first averaged over the ten feature-selection ratios. We then report
the mean and sample standard deviation over the ten random splits.
Higher values indicate better performance for AP and AUC, whereas lower
values are preferred for RL and ZL.

\section{Case Study}
\label{app:case}

This section presents a traceable example from SCENE.  SCENE is useful for
this purpose because it contains several independently confirmed conditions,
allowing the outputs of dataset analysis, triage, specialist consultation,
and objective construction to be followed as one continuous chain.  Numeric
values are rounded for presentation, while issue names, decisions, and
weights are copied from the persisted main-experiment artifacts.

\subsection{Dataset Analysis Trace}
\label{app:case_dataset_analysis}

The first case examines whether the generated issue card faithfully
summarizes the deterministic statistics.  The input contains both extremely
rare and extremely frequent labels, so the diagnosis can be checked directly
against observed counts rather than judged only from natural-language
plausibility.

\begin{casebox}{Case 1: SCENE Label-Distribution Analysis}
\small
\textbf{Analysis boundary.}
Only the \(3{,}080\) outer-training samples were profiled; none of the
\(1{,}320\) held-out samples contributed to the following statistics.

\textbf{Observed evidence.}
\begin{itemize}
    \item Number of analyzed samples: \(3{,}080\); number of features:
    \(634\); number of labels: \(33\).
    \item Minimum label frequency: \(0.000649\), corresponding to \(2\)
    positive samples.
    \item Maximum label frequency: \(0.968182\), corresponding to \(2{,}982\)
    positive samples.
    \item Label-imbalance ratio: \(1{,}491.0\); label-frequency Gini:
    \(0.6526\).
    \item Rare labels: \(13/33=39.39\%\); minimum positive count: \(2\).
\end{itemize}

\textbf{Triggered findings.}
\begin{itemize}
    \item \FHCode{EXTREME_LABEL_FREQUENCY}, severe;
    \item \FHCode{LABEL_IMBALANCE}, severe; and
    \item \FHCode{RARE_LABELS}, severe.
\end{itemize}

\textbf{Generated diagnosis.}
``Label distribution is severely skewed, with extreme rare/common labels, a
very large imbalance ratio, and many rare labels.''  The card further states
that positive frequencies vary severely and that the minimum positive count
is only two.
\end{casebox}

\paragraph{Analysis.}
The diagnosis is supported by several independent statistics: the frequency
extremes identify both tails, the imbalance ratio and Gini quantify global
skew, and the rare-label count measures how widespread the low-frequency
problem is.  The affected labels are also explicitly identified, including a
label with only two positives and a label present in more than \(96\%\) of
samples.  Consequently, the generated text neither invents an abnormality
nor exaggerates a weak signal; it is a concise rendering of the confirmed
training-only measurements.  The split manifest independently records the
training and test hashes, making the provenance of this profile auditable.

\subsection{Department Triage Trace}
\label{app:case_triage}

The second case tests whether related findings are routed to a coherent owner
without activating a separate specialist for every issue tag.  It also tests
whether a beneficial observation is kept as context instead of being treated
as a problem.

\begin{casebox}{Case 2: SCENE Evidence-to-Department Routing}
\small
\textbf{Activated department.}
\FHCode{LABEL_IMBALANCE}, with problem strength \(0.90\) and confidence
\(1.00\).

\textbf{Independent source issues.}
\begin{itemize}
    \item \FHCode{EXTREME_LABEL_FREQUENCY};
    \item \FHCode{LABEL_IMBALANCE}; and
    \item \FHCode{RARE_LABELS} and \FHCode{LOW_SAMPLE_PER_LABEL}.
\end{itemize}

\textbf{Absorbed findings.}
\begin{itemize}
    \item \FHCode{FEATURE_BUDGET_PRESSURE};
    \item \FHCode{WEAK_FEATURE_LABEL_SIGNAL_FOR_MINOR_}
    
    \FHCode{LABELS}; and
    \item \FHCode{LABEL_SIGNAL_IMBALANCE}.
\end{itemize}

\textbf{Supporting context.}
\FHCode{LARGE_LABEL_SPACE},
\FHCode{FEATURE_LABEL_SPECIFICITY_WEAKNESS},
\FHCode{DIFFUSE_RELEVANCE_SIGNAL}, and
\FHCode{WEAK_FEATURE_LABEL_SIGNAL}.

\textbf{Boundary decisions.}
\begin{itemize}
    \item \FHCode{FEATURE_QUALITY_DEFECT} is absorbed by
    \FHCode{VIEW_QUALITY_IMBALANCE}, because the observed sparsity and quality
    defects are explicitly view-specific.
    \item \FHCode{LOCAL_LABEL_INCONSISTENCY} is context-only because the
    profile confirms beneficial \FHCode{LOCAL_LABEL_CONSISTENCY}, not local
    neighborhood-label conflict.
\end{itemize}

\textbf{Routing reason.}
The extreme label-frequency skew is an independent treatment target.  Budget
pressure and weak minor-label signal compound rare-label handling and are
therefore assigned to the same owner instead of creating unrelated
treatments.
\end{casebox}

\paragraph{Analysis.}
Analysis. This routing preserves the distinction between independent
treatment targets and supporting evidence. The extreme label-frequency
skew is an independent problem, while feature-budget pressure and weak
minor-label signals describe its consequences in the context of limited
selection capacity. Absorbing these related findings into
LABEL\_IMBALANCE avoids introducing duplicated objectives while
preserving their evidence for the downstream Doctor.

For the quality-related findings, the triage decision reflects the
hierarchical relationship between view-level and feature-level issues.
Although quality degradation is observed, the evidence indicates that
the degradation originates from specific view characteristics, such as
view-level sparsity and heterogeneity, rather than isolated defective
features. Therefore, assigning these findings to VIEW\_QUALITY\_IMBALANCE
allows the downstream Doctor to address the underlying cause without
activating a separate feature-level treatment. Equally importantly, the
triage result does not activate LOCAL\_LABEL\_INCONSISTENCY from a finding
that confirms beneficial local consistency. The final routing is
therefore complete and non-redundant, and it was produced entirely from
issue cards computed on the outer-training partition.

\subsection{Specialist Diagnosis Trace}
\label{app:case_consultation}

The third case illustrates the Department Doctor's differential diagnosis.
The relevant question is not merely whether SCENE is multi-view, but which
implemented treatment matches the observed form of view heterogeneity.

\begin{casebox}{Case 3: SCENE View-Quality Consultation}
\small
\textbf{Department evidence.}
\begin{itemize}
    \item View zero rates:
    [0.00026,0.00231,0.41651,0.01198,
    
    0.00029].
    \item Zero-rate gap: \(0.4163\).
    \item View scale-heterogeneity gap: \(1.1055\).
    \item View feature counts: [64,\,225,\,144,\,73,\,128].
\end{itemize}

\textbf{Doctor diagnosis.}
Raw relevance scores are not directly comparable across views because the
views differ in sparsity, scale, and size.

\textbf{Selected treatment.}
\FHCode{view_normalized_relevance_}

\FHCode{loss}, with recommended weight \(0.30\),
moderate quality weighting, and score clipping.

\textbf{Treatments not selected.}
\begin{itemize}
    \item \FHCode{adaptive_view_allocation_loss}; and
    \item \FHCode{view_coverage_floor_loss}.
\end{itemize}

\textbf{Doctor rationale.}
Normalize relevance before cross-view competition, but do not force uniform
allocation or a hard view floor without evidence that such constraints are
needed.
\end{casebox}

\paragraph{Analysis.}
The selected loss directly addresses score comparability, which is the
mechanism indicated by the large sparsity and scale gaps.  A uniform
allocation term would answer a different question and could reserve features
for a weak view regardless of relevance.  Likewise, a floor would consume
top-\(k\) capacity without evidence of complete view exclusion.  Selecting
one calibration loss while rejecting the two stronger allocation treatments
shows that the Doctor used both positive and negative evidence rather than
mechanically enabling the entire Department catalog.  Both its profile and
prescription were derived from the same \(3{,}080\)-sample outer-training
partition.

\subsection{Pharmacist Fusion Trace}
\label{app:case_pharmacist}

The final case examines whether the Pharmacist turns overlapping local
prescriptions into a compact global objective.  In addition to the Doctors'
rationales, this decision used low-cost, training-only objective probes; no
classifier metric was provided to the Pharmacist.

\begin{casebox}{Case 4: SCENE Global Objective Construction}
\small
\textbf{Final enabled terms.}
\begin{itemize}
    \item \FHCode{label_weighted_relevance_loss}: primary, weight \(1.20\);
    \item \FHCode{redundant_cluster_quota_loss}: regularizer, weight \(0.12\);
    \item \FHCode{view_normalized_relevance_loss}: supporting, weight \(0.30\);
    \item \FHCode{negative_dependency_separation_loss}: guardrail, weight
    \(0.05\); and
    \item \FHCode{budget_penalty}: guardrail, weight \(0.2625\).
\end{itemize}

\textbf{Key disabled or merged terms.}
\begin{itemize}
    \item Graph-smoothed relevance was removed as a duplicate relevance
    reward and covered by the primary relevance backbone plus view
    calibration.
    \item Label-contribution balance was removed because its role was already
    covered by label-weighted relevance and its measured mechanical effect
    was weak.
\end{itemize}

\textbf{Low-cost decision clues.}
\begin{itemize}
    \item Removing cluster-quota redundancy damaged its target endpoint by
    \(0.5874\) and changed \(28.57\%\) of the top-\(k\) set.
    \item Removing label-contribution balance changed none of the top-\(k\)
    set and changed its target endpoint by only
    \(2.19\times10^{-5}\).
    \item Graph-smoothed and view-normalized relevance had gradient cosine
    \(0.9699\), indicating near-duplicate pressure.
    \item Raising the budget weight from \(0.15\) to \(0.2625\) improved its
    target endpoint by \(0.0237\), retained \(98.43\%\) top-\(k\) overlap,
    and incurred only a \(0.0052\) primary-endpoint cost.
\end{itemize}

\textbf{Fusion decision.}
Keep one label-balanced relevance backbone, retain the attributable
cluster-level redundancy treatment, preserve view normalization as the sole
secondary relevance-calibration term, and use separation and budget control
as guardrails.
\end{casebox}

\paragraph{Analysis.}
The final objective is smaller than the union of all Doctor prescriptions and
each retained term has a distinct role.  The Pharmacist keeps the redundancy
term whose removal materially changes both its endpoint and the selected
subset, while dropping label-contribution balance because it adds negligible
distinct pressure.  It also resolves duplicate relevance pressure by
retaining one primary backbone, keeping view normalization only as a
supporting term, and merging graph smoothing.  The budget dose is the only
upweighted term and is supported by an explicit local intervention rather
than a predictive test.  The mechanical probes used \(2{,}156\) samples
drawn solely from the outer-training partition and exposed no held-out labels
or classifier metric.  These actions are consistent with both the diagnosed
conditions and the local objective mechanics.  They support the
reasonableness of the fusion decision, although the probes themselves are not
claims of predictive optimality.

\newpage


\end{document}